\documentclass{article}

\usepackage{jfrExamplee}
\usepackage{soul} 
\usepackage{threeparttable}
\usepackage{tabularx}
\usepackage{arydshln}
\usepackage{array}
\usepackage{amsmath}
\usepackage{amssymb}
\usepackage{bm}
\usepackage[ruled, vlined, linesnumbered, commentsnumbered]{algorithm2e}
\usepackage[style=apa]{biblatex}
\usepackage{booktabs}
\usepackage{caption}
\usepackage{colortbl}
\usepackage{csquotes}
\usepackage{dsfont}
\usepackage{float}
\usepackage[symbol]{footmisc}
\usepackage{graphicx}
\usepackage{hhline}
\usepackage[bookmarks=true]{hyperref}
\usepackage{multicol}
\usepackage{multirow}
\usepackage[section]{placeins}
\usepackage{pgfplots}
\usepackage{setspace}
\usepackage{siunitx}
\usepackage{times}
\usepackage{upgreek}
\usepackage{varwidth}
\usepackage{verbatim}
\usepackage{xcolor}  
\usepackage[capitalise]{cleveref}

\DefineBibliographyStrings{american}{phdthesis = {PhD dissertation}}
\pgfplotsset{compat=newest}

\DeclareSIUnit{\belmilliwatt}{Bm}
\DeclareSIUnit{\dBm}{\deci\belmilliwatt}
\DeclareSIUnit{\sample}{S}

\newcommand{\ourMethod}[1]{Meta-Pilot}
\newcommand{\rssiOnly}[1]{Imp-RSSI}
\newcommand{\aoaOnly}[1]{cAoA(20 s)}
\newcommand{\aoaRssiA}[1]{AoA-RSSI(20 s)}
\newcommand{\aoaRssiB}[1]{AoA-RSSI(45 s)}

\providecommand{\keywords}[1]{
  \small	
  \textbf{\textit{Keywords---}} #1
}

\title{ConservationBots: Autonomous Aerial Robot for Fast Robust Wildlife Tracking in Complex Terrains}

\author{Fei Chen \\
School of Computer and Mathematical Sciences \\
The University of Adelaide \\
SA 5005, Australia \\
\texttt{fei.chen@adelaide.edu.au} \\
\And
Hoa Van Nguyen \\
Department of Electrical and Computer Engineering \\
Curtin University \\
WA 6102, Australia \\
\texttt{hoa.v.nguyen@curtin.edu.au} \\
\And
David A. Taggart \\
School of Animal and Veterinary Sciences \\
The University of Adelaide \\
SA 5005, Australia \\
\texttt{david.taggart@adelaide.edu.au} \\
\And
Katrina Falkner\\
School of Computer and Mathematical Sciences \\
The University of Adelaide \\
SA 5005, Australia \\
\texttt{katrina.falkner@adelaide.edu.au} \\
\And
S. Hamid Rezatofighi \\
Department of Data Science \& AI \\
Monash University\\
VIC 3800, Australia \\
\texttt{hamid.rezatofighi@monash.edu} \\
\And  
Damith C. Ranasinghe \\
School of Computer and Mathematical Sciences \\
The University of Adelaide \\
SA 5005, Australia \\
\texttt{damith.ranasinghe@adelaide.edu.au} \\
}

\begin{document}
\maketitle

\begin{abstract}
Radio tagging and tracking are fundamental to understanding the movements and habitats of wildlife in their natural environments. Today, the most widespread, \textit{widely applicable} technology for gathering data relies on experienced scientists armed with handheld radio telemetry equipment to locate low-power radio transmitters attached to wildlife from the ground. Although aerial robots can transform labor-intensive conservation tasks, the realization of autonomous systems for tackling task complexities under real-world conditions remains a challenge. We developed \textit{ConservationBots}---small aerial robots for tracking multiple, dynamic, radio-tagged wildlife. The aerial robot achieves robust localization performance and fast task completion times---significant for energy-limited aerial systems while avoiding close encounters with potential, counter-productive disturbances to wildlife. Our approach overcomes the technical and practical problems posed by combining a lightweight sensor with new concepts: i)~planning to determine both trajectory and measurement actions guided by an information-theoretic objective, which allows the robot to strategically select near-instantaneous range-only measurements to achieve faster localization, and time-consuming sensor rotation actions to acquire bearing measurements and achieve robust tracking performance; ii)~a bearing detector more robust to noise and iii)~a tracking algorithm formulation robust to missed and false detections experienced in real-world conditions. We conducted extensive studies: simulations built upon complex signal propagation over high-resolution elevation data on diverse geographical terrains; field testing; studies with wombats (\textit{Lasiorhinus latifrons}; nocturnal, vulnerable species dwelling in underground warrens) and tracking comparisons with a highly experienced biologist to validate the effectiveness of our aerial robot and demonstrate the significant advantages over the manual method.
\end{abstract}


\section{Introduction}
Understanding movements, activities, and patterns in animal behaviors are essential for biodiversity conservation, natural resource management, and precision agriculture.
Today, field scientists employ multiple techniques, such as vision-based sensors  \parencite{selby2011autonomous, christ2014, olivares2015towards, gonzales2016, arunabha2023}, Global Positioning System (GPS) tags, or Very High Frequency (VHF) tags \parencite{cochran1963radio,kenward2000manual, foley2020, jin2023} to study animal behaviors, movements, and activity. 
Despite the advances in technology, VHF radio telemetry or radio-tracking is still the most important tool employed to study the movement of animals in their natural environments. 
Because they are smaller, lightweight, suited for nearly all mammal and bird species, and operate for longer in comparison to GPS-based counterparts. Consequently, remain a popular and cost-effective technique for field studies \parencite{VHF_bird_on_the_move, wikelski2007going, saunder2022}. 
However, the traditional method of radio-tracking typically requires researchers to trek long distances in the field, armed with cumbersome radio receivers with hand-held antennas and battery packs to manually home in on radio signals emitted from radio-tagged or collared animals. 
The precious spatial data acquired through radio-tracking comes at a significant cost to researchers in terms of manpower, time, and funding. The problem is often compounded by other challenges, such as low animal recapture rates, equipment failures, and the inability to track animals that move into inaccessible terrain or underground burrows.

Developments in low-cost unmanned aerial vehicles (UAVs) with the capacity to carry payloads, such as radio receivers and antennas \parencite{anderson2017drones, lahoz2021}, is a potential solution.
Because of the advantages offered by the ease of deployment and \textit{high mobility}, UAVs have the potential to automate and scale up manual tasks to significantly reduce the time, labor, and cost of employing traditional tracking approaches.
Early achievements in \textit{autonomous systems} for wildlife tracking  have demonstrated robotic platforms for the task \parencite{tokekar2010, vander2014, Cliff2018, hoa2019jofr, koray2023}. The approaches localize VHF radio-tagged animals using either the Receiver Signal Strength Indicator (RSSI) or Angle of Arrival (AoA) of radio signals emitted from radio tags where the robot's trajectory planning algorithm endows autonomy to improve the localization accuracy---\parencite{Cliff2015, Cliff2018} or reduce the tracking error--- \parencite{hoa2019jofr}. Despite the recent advancements, the realization of an autonomous UAV capable of dealing with the technical and practical complexities of the problem remains a challenge. For example: 
\begin{itemize}
    \item RSSI-based approach---capable of rapid measurements and benefiting from a simple receiver that requires a single directional antenna, only demonstrates superior performance across mostly flat terrains or when the radio propagation model is accurately known \parencite{hoa2019jofr}. But, building and employing an accurate radio propagation model requires access to terrain information and dealing with variables that can change dynamically, such as changes in signal attenuation due to the appearance of trees and the impact of moisture conditions on signal propagation. 
    \item AoA approach---although more robust in unfamiliar or complex environments, requires a larger and bulky antenna array and long measurement acquisition times; 45 seconds per measurement \parencite{Cliff2015}.
    \item Practically, autonomous systems need to operate under limited battery power, on-board processing, flight times, and payload capabilities of UAVs (ideally, an aerial robot should fly, track and locate animals and return to base without needing an intervening battery change away from the base).
    \item Multiplicity of VHF radio signal propagation in complex terrains complicates automatic localization algorithms and detection of, often weak signals, of simplistic wave modulations from VHF radio collars.
    \item Potential disturbances to animals caused by the operation of the UAV \parencite{uav_disturb1, HODGSON2016R404, uav_disturb3}, is counterproductive to the task and requires mitigation strategies. 
\end{itemize}
 
Our work formulates a planning problem for a new \textit{hybrid} approach to exploit the simple and fast RSSI measurement acquisitions and selectively exploit the slower, more robust AoA measurements by providing an aerial robot the autonomy to plan not only its trajectory to track and locate animals but also its measurements to track multiple, mobile, radio-tagged wildlife, simultaneously. The robot we have developed is fast, robust, and scalable to simultaneously track and localize multiple radio-tagged wildlife while planned trajectories \textit{minimize} disturbances to wildlife. We summarize our main contributions below:

\begin{figure}[hbt!]
    \centering
    \includegraphics[width=0.8\textwidth]{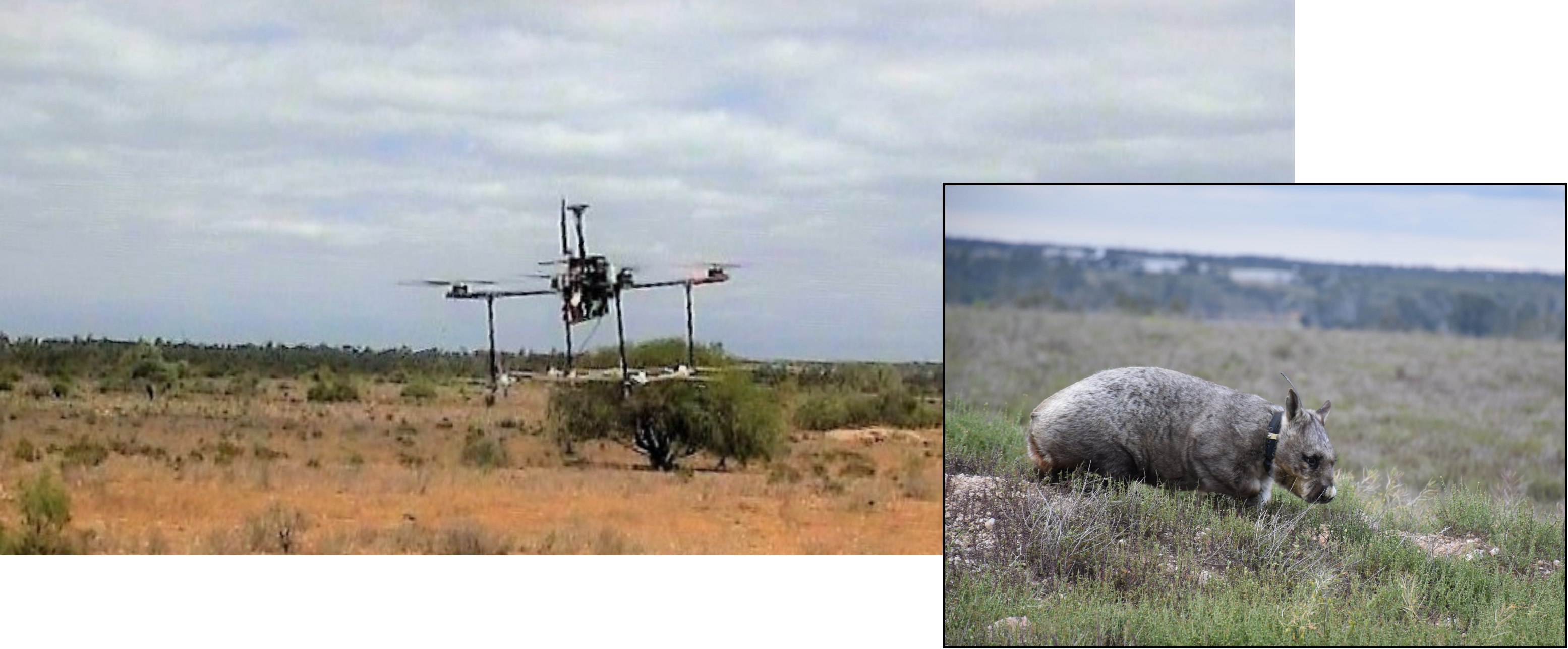}
    \caption{A ConservationBot in flight, right after take-off to track and locate Southern Hairy-nosed wombats. Inset: Southern Hairy-nosed wombat \emph{Lasiorhinus latifrons} released into their habitat after being tagged with a VHF radio collar.}
    \label{fig:first_fig}    
\end{figure}

\begin{enumerate}
    \item We propose planning not only trajectories but also the measurement method. The \textit{planning} algorithm formulation determines the most informative trajectory and measurement acquisition actions to reduce the time needed to track and locate multiple, dynamic, radio-collared wildlife in various terrain conditions.
    The planner allows the robot to i)~exploit the simplicity and rapidity of RSSI measurement acquisitions for range-only tracking; and ii)~the time-consuming bearing measurement method when range-only measurement uncertainty is high. Importantly, to avoid close approaches to wildlife and minimize potential disturbances, trajectory planning is constrained by a probabilistic void region. 
    
    \item Our state estimation algorithm for tracking is robust across variable environmental and terrain conditions  as well as VHF signal propagation artifacts to localize radio-tagged wildlife in various outdoor  environments in 3D settings without detailed terrain information. To achieve this robustness: i)~we integrate an imprecise likelihood function to account for the complex radio propagation effects such as signal diffraction and vegetation attenuation, removing the need for a precise RSSI measurement model. This method expands the versatility of the estimation algorithm to allow RSSI measurements to be used for tracking when an accurate model of measurements in a given terrain is difficult to build; 
    ii)~we propose a bearing detector based on the rotation AoA measurement method to generate more robust AoA measurements under noisy conditions; and iii)~compared to prior approaches, we have \textit{explicitly} considered practical challenges in \textit{tracking} radio-tagged animals such as missed measurements, object birth, and death, poor signal-to-noise ratio impacting signal detection probabilities and formulated a Bernoulli Bayesian filter (BF) for the tracking task.
    
    \item To examine the performance of our system, we used extensive Monte-Carlo simulations modeling complex VHF signal propagation over 3D terrains with different levels of complexity and extensive field experiments.
    Our field testing included over $30$ missions and a pilot study with radio-tagged southern hairy-nosed wombats (\textit{Lasiorhinus latifrons}), a nocturnal, burrowing species that uses underground warrens and, therefore, challenging to track with manual methods as well as an autonomous system \textit{versus} expert-human-tracker field trials to demonstrates the efficacy, performance, and versatility of our system.
\end{enumerate}

The rest of this paper is organized as follows. Section \ref{sec:relative_works} presents related studies in radio tracking and localization methods employing unmanned aerial vehicles along with methods for estimating the state of objects, such as radio-tagged wildlife using noisy measurements. In Section \ref{sec:proposed_method} we present the problem formulation, the proposed methods and system underpinning our fast, robust aerial robot for locating multiple wildlife in complex terrains. Section~\ref{sec:simulation_experiment} evaluates our proposed methods through an extensive series of simulation-based scenarios. Our proposed robot prototype construction is described in Section \ref{sec:prototype}. Section \ref{sec:fieldexp} describes the series of field experiments and results confirming the effectiveness of our approach followed by lessons learned in Section \ref{sec:lessons_learned} and concluding remarks in Section~\ref{sec:conclusion}.

\section{Related Work} \label{sec:relative_works}
The problem of using UAVs to localize radio sources has been studied recently in the literature. More broadly, existing studies can be categorized based on the measurement methods employed; those using RSSI-based methods or those using AoA-based methods. 
The design of the system and algorithms are primarily a function of the measurement method employed. Hence it is useful to consider previous methods from this perspective. We also provide a brief review of related works in Multi-object tracking methods since our work focuses on tracking multiple animals.

\subsection{AoA-Based Systems}
A widely used method for measuring the AoA of RF signal is through the use of a phased array antenna. While the method can measure AoA with high accuracy with minimum measurement time, it requires specialized hardware and sophisticated signal processing algorithm. Therefore, such a sensor payload is difficult to mount and employ on a UAV platform due to weight, size, and processing power constraints.
Consequently, an alternative approach has emerged using a directional antenna, which is rotated to determine the direction of the signal source to detect AoA. \cite{ground_rotate_ant}, an early study, demonstrated a ground-based robot system that used a rotating directional antenna to determine the AoA and locate the source of a wireless node. \cite{monowing_aoa} later developed an RF source-seeking system with a single-wing rotating micro aerial vehicle. By fitting a directional antenna to its wing and exploiting the natural rotation of the vehicle, the system can quickly estimate the AoA of an RF source at each rotation.

Early efforts to demonstrate the rotating AoA-based methods for radio-tagged animal localization were reported in \cite{tokekar2010},\cite{Cliff2015},\cite{VonEhr_SDR_AoA2016}, \cite{Cliff2018}, and \cite{UAV_RT}. 
The studies developed a multi-rotor UAV system with a path-planning algorithm to direct the UAV to collect AoA measurements. AoA methods are robust against the multi-path effects of radio propagation but the measurement time required (45 seconds per measurement in \cite{Cliff2015}) is significantly longer than RSSI-based approaches with near-instantaneous measurements. The impact of longer measurement times is more practical---it limits the maximum search area, increases flight times to complete a task, and reduces the ability to locate mobile animals. For example, while undertaking a slow rotation can increase the accuracy of the acquired AoA measurement for static wildlife, it is counterproductive if objects are mobile. 

An alternative, a pseudo-bearing approach \parencite{Dressel_pseudo_bearing}, sought to address the limitations of rotational AoA methods by incorporating an additional uni-directional antenna along with a directional antenna, albeit for operation at a much higher frequency, 2.4~GHz, than VHF. As a result, the methods can perform measurement updates more quickly and improve localization time.
However, this approach requires a more complex radio receiver with multiple antennas; consequently, it increases the weight of sensor payload on a UAV, especially in VHF signal tracking scenarios necessitating antennas with large physical dimensions, while trajectory planning with such an approach also remains to be demonstrated in practice.

\subsection{RSSI-Based Systems}
In contrast, the RSSI-based systems utilize signal strength to estimate the distance between the radio transmitter and the receiver. This approach only requires a simple and lightweight receiver and antenna. 
The use of RSSI-based measurements on board a UAV to locate radio-tagged wildlife was demonstrated by \parencite{5654385}, where a fixed-wing UAV equipped with a directional antenna was used to locate a fixed location radio tag. Then, a system based on a multirotor UAV with an omnidirectional antenna was presented in \cite{santos2014}. The approach employed a receiver to capture measurements of the radio signal's signal-to-noise ratio and estimated the radio tag's position \textit{offline}. \cite{hoa2019jofr, LAVAPilot} demonstrated an RSSI-based aerial robot with online path planning for RSSI measurements and a particle filter-based estimation method using a customized, lightweight directional antenna for localizing multiple \textit{mobile} objects on relatively flat terrains. \cite{small_drone_rssi} took a similar approach using a small UAV with a dipole antenna to collect RSSI measurements along a fixed trajectory where the positions of radio-collared animals were determined \textit{offline} based on a signal propagation model. 

In contrast to AoA-based methods for the task, an RSSI-based system is more efficient due to its simpler receiver and faster measurement acquisition times. However, the key limitation is sensitivity to environmental effects impacting radio signal propagation, signal diffraction, scattering, and vegetation attenuation. Because propagation characteristics of the radio signal need to be accurate but, in practice, can be difficult to model. Topographical variations, vegetation coverage, or weather can result in unpredictable attenuation of radio signals and thus limit the scenarios in which RSSI-based methods can be reliably applied. As a consequence, RSSI-based methods are mostly used in less complex environments where radio propagation is easily modeled and predictable. 
    
\subsection{Multi-Object Tracking} 
The primary problem in multi-object tracking is to estimate the state of multiple objects when the associations between measurements and objects are unknown. Traditional methods, including joint probabilistic data association (JPDA) filter \parencite{bar1987tracking}, multiple hypothesis tracking (MHT) filter \parencite{blackman1987} explicitly associate measurements and objects. More recent approaches based on random set statistics\cite{10.5555/1512927} have led to methods such as the probability hypothesis density (PHD) filter\parencite{mahler_phd}, cardinalized PHD (CPHD) filter \parencite{mahler_cphd}, multiobject multi-Bernoulli filter \parencite{10.5555/1512927, vo_cardinality_2009}, generalized labeled multi-Bernoulli (GLMB) filter \parencite{vo_labeled_2013} and labeled multi-Bernoulli (LMB) filter \parencite{reuter_labeled_nodate}.
However, in our problem, individual radio-collared wildlife can be uniquely identified by the frequency of its radio-collar signal. Therefore, we do not need to solve the complex data association problem.
The model for measurements (RSSI or AoA) is non-linear, therefore a filter suitable for non-linear systems, such as a particle filter \parencite{gordon_novel_1993} was used in prior robotic systems for the task. In contrast, we consider a particle implementation of a Bernoulli filter \parencite{10.5555/1512927} formulation not only to account for a non-linear system but also to explicitly model practical signal propagation effects, such as measurement miss detections and false detections into the formulation. Thus, making the method of estimating the location of wildlife more robust than the particle filter.

\subsection{Summary}
RSSI-based and AoA-based methods are commonly used for estimating the location of radio sources. When radio propagation can be accurately modeled, the RSSI-based methods provide significant advantages over AoA-based approaches, given their simplistic receiver design and low measurement time. However, in a complex environment, the AoA approach is a more robust method due to it being invariant to various environmental variables and the difficulty of building accurate propagation models for complex terrains to support RSSI-based methods. We present an approach that combines the advantages of these two measurement approaches; an aerial robot system that takes advantage of both methods while minimizing their limitations.

Importantly, existing systems, irrespective of the measurement method, used an online estimator to determine the location of objects. The estimator, based on Bayesian estimation theory, requires  accurate noise models and sensor measurement models to determine the probability distribution of objects; these estimation methods include particle filters \parencite{hoa2019jofr}  \parencite{5654385}, grid filters \parencite{Cliff2015}\parencite{Cliff2018} and Kalman filters \parencite{jensen2013}. Notably, these estimators cannot handle practical challenges with trackings, such as missed detections and false detections; explicitly modeling these real-world conditions may lead to better estimation accuracy.

\begin{figure}[hbt!]
    \centering
    \includegraphics[width=0.9\columnwidth]{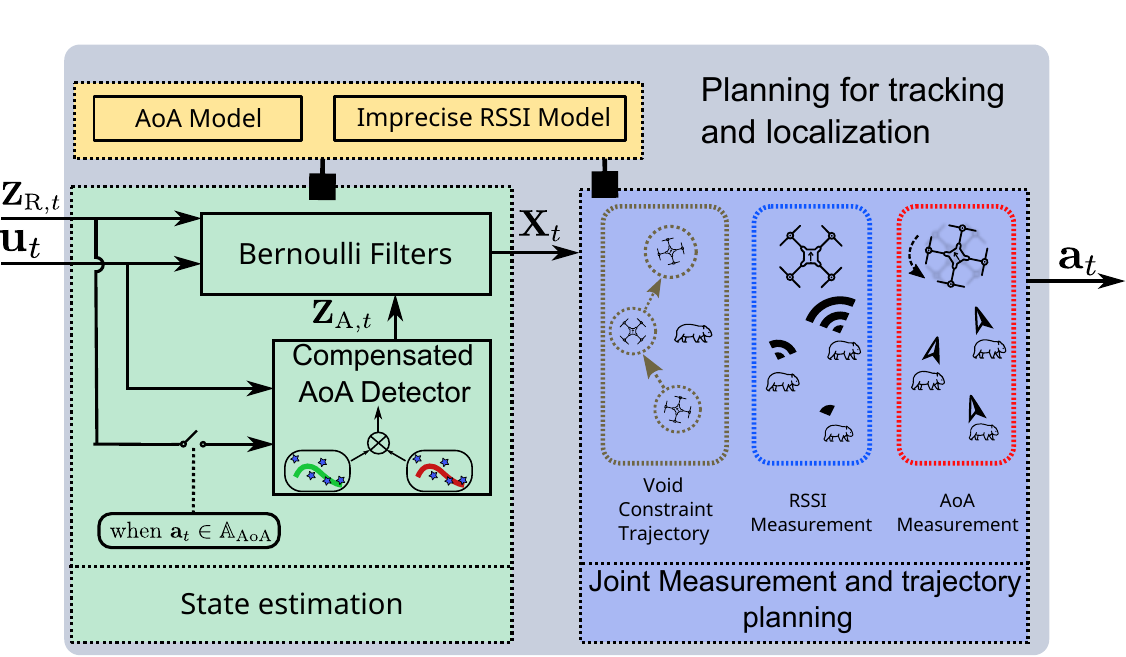}
    \caption{An overview of the proposed Bayesian-POMDP theoretical framework to realize an autonomous aerial vehicle for fast, robust tracking of multiple wildlife in complex terrains. Here, the UAV state is denoted by $\mathbf{u}_{t}$, and the belief densities of the set of objects (in our case wildlife) are denoted by $\mathbf{X}_{t}$. Briefly: i)~the proposed \textit{compensated AoA measurement detector} employs RSSI measurements $\mathbf{Z}_{\text{R}, t}$ during an AoA measurement action $\mathbf{a}_t \in \mathbb{A}_{\text{AoA}}$ to generate an AoA measurement $\mathbf{Z}_{\text{A}, t}$; ii)~Bernoulli filters utilize RSSI measurements $\mathbf{Z}_{\text{R}, t}$ along with the imprecise RSSI model and AoA measurements $\mathbf{Z}_{\text{A}, t}$ along with the AoA model at time $t$ to achieve robust estimations of object states (e.g., the position of each wildlife); and iii)~the \textit{new} measurement (AoA, RSSI) and trajectory planning formulation using a POMDP generate control actions $\mathbf{a}_{t}$ while ensuring the UAV maintains a safe distance from the wildlife of interest by generating void constrained trajectories.}
    \label{fig:planning_block}    
\end{figure}

\section{Proposed Planning for Tracking and Localization Problem Formulation} \label{sec:proposed_method}
We consider the problem of controlling a UAV equipped with a simple sensor system---a directional antenna and a digital signal processing module---for autonomously localizing multiple radio-tagged wildlife while maintaining a safe distance from the wildlife of interest to prevent potential disturbances. Performing tracking (estimating the positions of individual radio-tagged wildlife over time) in real-time necessitates an online estimation method. And performing the task autonomously necessitates a dynamic planning method for robot navigation. In contrast to previous problem formulations, we consider incorporating an RSSI measurement model uncertainty to remove the need for accurate measurement models---difficult to derive in practice due to changing terrain and environmental conditions---and consider practical signal detection artifacts such as missed and false detection to formulate a robust method of estimating the locations of wildlife using a Bernoulli Bayesian filter. Further, in contrast to previous approaches, we consider dynamically planning both the trajectory and the signal measurement method using a POMDP (partially observable Markov decision process) formulation to allow the autonomous selection of the most informative measurement method: i)~simple and fast RSSI measurements; or ii)~slower but more robust AoA measurements. 

Figure~\ref{fig:planning_block} provides an overview of our proposed planning for tracking and localization approach  built upon a joint Bayesian-POMDP theoretical framework, which includes: 
i)~an AoA measurement model and an RSSI measurement imprecise model for increased tracking and localization robustness against the impacts of varying terrain and environmental conditions; 
ii)~compensated rotation AoA measurement method to generate AoA measurements with higher accuracy under noisy conditions; 
iii)~Bernoulli Filter employing both RSSI and AoA measurements to produce estimated object states (tracks), even under low measurement detection probabilities, experienced in \textit{practical} system deployment settings; iv)~measurement and trajectory planning to select both the best trajectory and measurement method for faster and more robust localization under different terrain conditions while \textit{minimizing disturbances} to the wildlife of interest.  In the following sections, we detail our formulation of real-time planning for tracking and localizing wildlife described in  Figure~\ref{fig:planning_block}. 

\subsection{State Estimation and Measurement Models}
This section presents our online tracking and localizing formulation under the theoretical framework of a Bernoulli Bayesian filter to formulate a robust method of estimating the locations of wildlife using the proposed AoA measurement detection method, the associated measurement model, and the imprecise RSSI measurement model. 

Prior to proceeding further, we introduce the following conventions for notation consistency: standard letters (\emph{e.g.} $x,~X$) for scalar values, lowercase bold letters (\emph{e.g.} $\mathbf{x}$) for vector values (\emph{e.g.} single-object states); bold capital letters (\emph{e.g.} $\mathbf{X}$) for set values (\emph{e.g.} multi-object states); blackboard letters (\emph{e.g.} $\mathbb{X}$) for state spaces.

\subsubsection{Radio Signal Model of a VHF Wildlife Collar}
It is first useful to understand the nature of the signal source since the measures of this signal will need to be employed to estimate the location of each wildlife. Each radio tag employed for studying wildlife emits an on-off-keying signal at a unique frequency $f$ with an unknown time offset $\tau\in\mathbb{R}_{0}^{+}$ as illustrated in Figure~\ref{fig:raw_signal}. 
Now, let's denote the state of the UAV (observer) as $\mathbf{u} = [\mathbf{u}_{l}^{T}, \theta^{(u)}]^{T} \in  \mathbb{R}^{3} \times [0,2\pi)$, including its position $\mathbf{u}_{l}=[u_{x}, u_{y}, u_{z}]^{T}$ and heading angle $\theta^{(u)}$ and the state of each radio-tagged wildlife as $\mathbf{x}=[p_{x}, p_{y}, p_{z}]^{T}\in \mathbb{X}\subseteq\mathbb{R}^{3}$, where $(\cdot)^{T}$ is the matrix transpose. 
Then, the noiseless signal $\chi(t)$  at time $t$ from an object with state $\mathbf{x}$ received by a directional antenna mounted on a UAV with state $\mathbf{u}$ in the far field region can be modeled as \parencite{hoa2019tsp}: 

\begin{figure}[hbt!]
    \centering
    \includegraphics[width=0.5\textwidth]{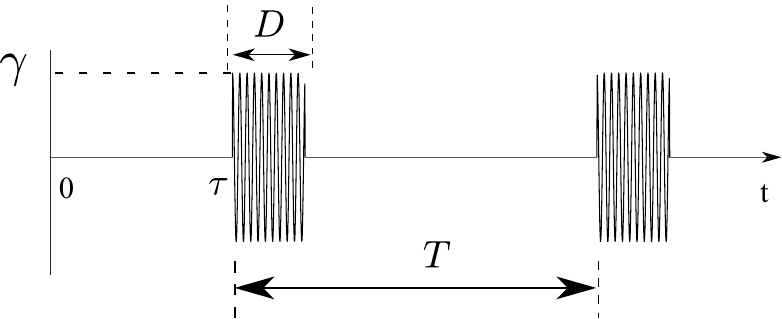}
    \caption{On-off-keying signal with pulse width $D$, period $T$, time offset $\tau$ and amplitude $\gamma$.}
    \label{fig:raw_signal}    
\end{figure}

\begin{equation}
    \chi(t) = \gamma(\mathbf{x}, \mathbf{u})\cos[2\pi ft + \Phi_{t}]\operatorname{rect}(D,T,t - \tau)
    \label{eq:raw_signal_eqn}
\end{equation}
where
\begin{itemize}
    \item $\Phi_{t}\in\mathbb{R}$ is the received signal phase at time $t$;
    \item $\gamma(\mathbf{x}, \mathbf{u}) = \Gamma(d_{0})G_{r}G_{a}(\zeta(\mathbf{x}, \mathbf{u}))(d_{0}/d(\mathbf{x, u}))^{n/2}$ is the received signal amplitude when the distance between object $\mathbf{x}$ and observer $\mathbf{u}$ is $d(\mathbf{x, u})$ and with source signal amplitude $\Gamma(d_{0})$, which is measured at reference distance $d_{0}$ \parencite{1458287};
     $n\in\mathbb{R}^{+}$ is the environment-dependent path-loss exponent; 
    $G_{r}$ is the receiver gain; 
    $G_{a}(\cdot)$ is antenna gain pattern;
    $\zeta(\mathbf{x, u}) = \operatorname{arctan2}(\frac{p_{x} - u_{x}}{p_{y} - u_{y}})-\theta^{(u)}$ which convert the azimuth angle from object $\mathbf{x}$ to observer $\mathbf{u}$ into the local reference frame of the observer with heading $\theta_{u}$;
    
    \item $\operatorname{rect}(D,T,\cdot)$ is a periodic rectangular function with pulse width $D$ and period $T$, and $T > D$:
    \begin{equation}
        \operatorname{rect}(D,T,t - \tau) = \sum_{i = -\infty}^{\infty}\operatorname{boxcar}(\tau,\tau+D,t + iT)
    \end{equation}
    and 
    \begin{equation}
        \operatorname{boxcar}(a,b,x) = \begin{cases}
            1, &a \leq x \leq b \\
            0, & \text{otherwise}
        \end{cases}
    \end{equation}
\end{itemize}

\subsubsection{Bernoulli Filter}
To infer the unknown state $\mathbf{x}\in \mathbb{X}$ (3D coordinates of wildlife) given noisy measurements $z\in\mathbb{Z}$, AoA and RSSI measurements extracted from the received signals, we consider a Bernoulli filter---also known as JoTT or joint object detection and tracking filter---\parencite{10.5555/1512927}\parencite{Alsolami2012Auth}\parencite{6178069}. Recall each object---radio-collared wildlife---emits a signal at a unique frequency; hence a unique object's state $\mathbf{x}$ can be estimated from the measurement and tracked independently. Thus, we do not need to solve the complex data association problems typical of a multi-object tracking setting and can estimate the state of each object using a Bernoulli filter formulation independently. 

The Bernoulli filter is an exact Bayesian filter based on the random finite set theory (RFS) \parencite{10.5555/1512927}. 
Notably, the filter is capable of handling practical signal detection artifacts such as missed and false detections and dealing with the reality of animals with VHF radio collars wandering in or out of the sensor's detection range in a unified framework.

The Bernoulli RFS $X$ can either have at most one element with probability $r$ distributed over the state space $\mathbb{X}$ according to probability density function (PDF) $s(\mathbf{x})$ or empty with probability $1 - r$:
\begin{equation}
    \Psi(\mathbf{X}) = \begin{cases} 1-r, & \text{if } \mathbf{X} = \emptyset \\ r\cdot s(\mathbf{x}), & \text{if } \mathbf{X} = \{\mathbf{x} \} \\ 0 & \text{if } |\mathbf{X}| \geq 2 \end{cases}
    \label{eq:BF_RFS}
\end{equation}
where $|\mathbf{X}|$ denotes the cardinality of $\mathbf{X}$.

Given measurement set $\mathbf{Z}_{t} = \{z^{(1)}_t, \ldots, z^{(|\mathbf{Z}_{t}|)}_t \}$ at time $t$,
the posterior distribution $\Psi(\mathbf{X}_{t}|\mathbf{Z}_{1:t})$ from time $t-1$ to time $t$ can be propagated in two steps, the \textit{prediction} step and \textit{update} step. 
Notably, Bernoulli RFS \eqref{eq:BF_RFS} is entirely described by its existence probability $r$ and single object PDF $s(\mathbf{x})$. 
Therefore, the prediction and update step of \eqref{eq:BF_RFS} only needs to propagate $r$ and $q_{\mathbf{x}}$.

The prediction step for the Bernoulli filter is
\begin{align}
    \begin{split}
        r_{t|t-1} &= r_{b}\cdot(1-r_{t-1|t-1}) + r_{s}\cdot r_{t-1|t-1} \\
        q_{t|t-1}(\mathbf{x}) &= \frac{r_{b}\cdot(1-r_{t-1|t-1})\cdot b_{t|t-1}(\mathbf{x}) + r_{s}\cdot r_{t-1|t-1}\cdot\int q_{t|t-1}(\mathbf{x|x^{'}})\cdot s_{t-1|t-1}(\mathbf{x^{'}})d\mathbf{x^{'}}}{r_{b}\cdot(1-r_{t-1|t-1})+r_{s}r_{t-1|t-1}}
    \end{split}
    \label{eq:bf_predict}
\end{align}
$r_{b}$ is the probability of object birth and $b_{t|t-1}(\mathbf{x})$ is the spatial distribution of predicted object birth.
These two parameters models object to enter or leave a search space.
In our context, wildlife may disappear from a search space, such as going underground or appearing suddenly from burrows, as is the case with the species we investigate in our field trials.
$q_{t|t-1}(\mathbf{x|x^{'}})$ is the object transitional density, which describes the object's dynamic.

The update step for the Bernoulli filter is
\begin{align}
    \begin{split}
        r_{t|t} &= \frac{1 - \Delta_{t}}{1 - \Delta_{t}\cdot r_{t|t-1}} \\
        q_{t|t}(\mathbf{x}) &= \frac{1 - P_{D}(\mathbf{x}) + P_{D}(\mathbf{x})\sum_{\mathbf{z}\in\mathbf{Z}_{t}}\frac{L_{t}(\mathbf{z|x})}{\lambda c(\mathbf{z})}}{1-\Delta_{t}}s_{t|t-1}(\mathbf{x})
    \end{split}
    \label{eq:bf_update}
\end{align}

where
\begin{equation}
    \Delta_{t} = \int P_{D}(\mathbf{x})s_{t|t-1}(\mathbf{x})d\mathbf{x} - \sum_{\mathbf{z}\in\mathbf{Z}_{t}}\frac{\int L_{t}(\mathbf{z|x})s_{t|t-1}(\mathbf{x})d\mathbf{x}}{\lambda c(\mathbf{z})}
\end{equation}

with $L_{t}(\mathbf{z|x})$ being the measurement likelihood function and $\lambda$ being the expected number of false measurement with PDF $c(\mathbf{z})$.
$P_{D}(\mathbf{x})$ is the probability of detection given state $\mathbf{x}$.

For more detailed derivation and implementations of the Bernoulli filter, we refer the reader to \parencite{10.5555/1512927}\parencite{6497685} for further reference.

The filter update step \eqref{eq:bf_update} requires the likelihood of measurements $L_{t}(\mathbf{z|x})$ to obtain the posterior distribution. We derive the measurement likelihood for two types of measurements; i)~RSSI; and ii)~AoA. Recall we consider an imprecise measurement model for RSSI; we describe its formulation next, followed by our proposed robust AoA detector formulation and the AoA measurement model for filter updates.

\subsubsection{Imprecise RSSI Measurement Model} \label{sec:imprecise-rssi-model}
Given signal with form \eqref{eq:raw_signal_eqn}, the primary measurement can be obtained is RSSI measurement, which is completely characterized by $\gamma(\mathbf{x, u})$ and defined as its Root-Mean-Square (RMS) power.
Suppose the receiver gains $G_{r}=1$, then an RSSI measurement can be expressed as:

\begin{align}
\begin{split}
    z_R =  h_{R}(\mathbf{x, u}) & = 10\log_{10}\big((\gamma(\mathbf{x}, \mathbf{u})/\sqrt{2})^{2}\big)\\
    & =\tilde{\Gamma}(d_{0}) - 10n\log_{10}(d(\mathbf{x, u})/d_{0})+\tilde{G}_{a}(\zeta(\mathbf{x, u}))
    \label{eq:ideal_rssi_meas}
\end{split}
\end{align}
where $\tilde{\Gamma}(d_{0}) = 10\log_{10}((\frac{\Gamma(d_{0})}{\sqrt{2}})^{2})$,  and $\tilde{G}_a(\zeta(\mathbf{x,u})) = 10\log_{10}(G_a(\zeta(\mathbf{x,u}))^{2})$.

In a non-urban environment, the received radio signal is usually corrupted by environmental noise and can be modeled as
\begin{equation}
    z_{R} = h_{R}(\mathbf{x, u}) + \mathit{w}_{R}  \label{eq:RSSI_meas}
\end{equation}
where $\mathit{w}_{R}\sim \mathcal{N}(\cdot; 0,\sigma^2_R)$ is measurement Gaussian white noise with covariance $\sigma_R^2$.

\eqref{eq:RSSI_meas} yields the RSSI likelihood function:
\begin{equation}
    \mathcal{L}_{R}(z_{R}; \mathbf{x, u}) = \mathcal{N}(z_{R}; h_{R}(\mathbf{x, u}),  \sigma_{R}^{2})
    \label{eq:range_likelihood}
\end{equation}
where $\mathcal{N}(\cdot;\mu, \sigma^{2})$ is the Gaussian probability density function with mean $\mu$ and variance $\sigma^{2}$.

\begin{figure}[hbt!]
    \centering
    \includegraphics[width=0.8\textwidth]{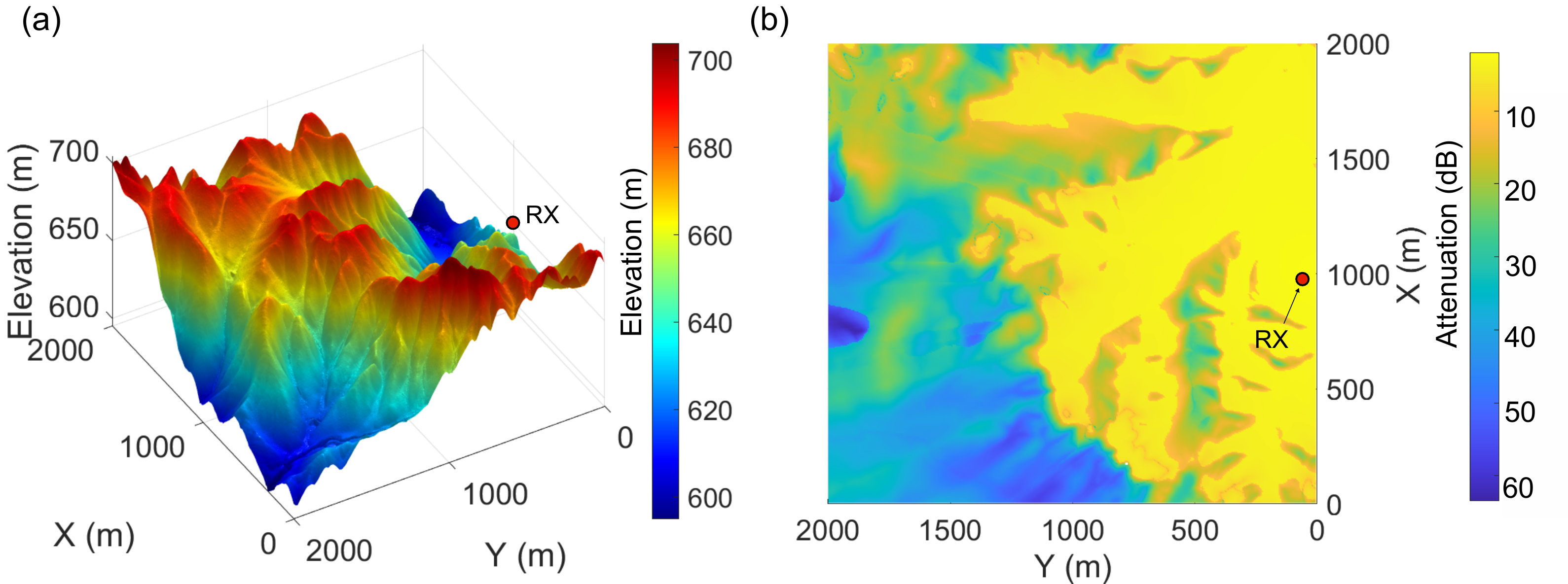}
    \caption{(a) An example complex terrain from  \parencite{DEM_australia} with a fixed receiver (RX) marked by a red circle; (b) Simulated environment-dependent signal strength attenuation resulting \textit{only} from terrain loss and vegetation loss, without the attenuation component $(d_{0}/d(\mathbf{x, u}))^{n}$ over distance with a transmitter placed at each coordinate point on the terrain map. For transmitter locations without significant blockage from terrain conditions (locating at $Y < \SI{1000}{\meter}$), the signal attenuation is significantly less than those blocked by the terrain (most locations when $Y > \SI{1000}{\meter}$. Notably, terrain loss simulated is formulated in \eqref{eq:terrain_loss} and vegetation loss is formulated in \eqref{eq:vegetation_loss} in Section~\ref{sec:sim_model}.}
    \label{fig:loss_illu_sim}    
\end{figure}

The path-loss model in \eqref{eq:RSSI_meas} is accurate when the receiver is within direct line-of-sight to the transmitter and other forms of loss are negligible.
However, in complex terrain conditions, other forms of loss introduced by multi-path propagation, diffraction, scattering, shadowing, or attenuation due to vegetation, are not negligible, which makes such a model generally inadequate. An illustration of vegetation and terrain loss variations over an example terrain gathered from digital elevation map data from  \parencite{DEM_australia} is presented in Figure.~\ref{fig:loss_illu_sim}. Without detailed terrain and site information (such as vegetation), it is generally difficult or impractical to construct an accurate measurement model, especially under complex terrain conditions such as hills, mountains, and varying vegetation conditions.

In order to handle the practical constraints of using RSSI measurements, we consider incorporating a model uncertainty or \textit{imprecision} to remove the need for an accurate measurement model. 
We introduce an additional error term $\mu_{S}(\textbf{x, u})$ which can represent any practical propagation complexities that can cause the RSSI measurements to deviate from the simple model in \eqref{eq:RSSI_meas}. 
Now, we express the RSSI measurement model that incorporates various measurement model uncertainties:
\begin{equation}
    z_{R} = h_{R}(\mathbf{x, u}) + \mu_{S}(\mathbf{x, u}) + \mathit{w}_{R}.  \label{eq:range_meas2}
\end{equation}

The term $\mu_{S}(\mathbf{x, u})$ can be considered as an unknown parameter of the measurement function, and \eqref{eq:range_meas2} can be rewritten as:
\begin{equation}
    h(\mathbf{x, u}; \mu) 
    = h_{R}(\mathbf{x, u}) + \mu_{S}(\mathbf{x, u}) 
    \label{eq:imprecise_function}
\end{equation}
where we refer $\mu\in[\mu_{min}, \mu_{max}]$ as the  \textit{(RSSI) model imprecision},  and $\mu_{min}=\min(\mu_{S}(\mathbf{x, u}))$ and $\mu_{max}=\max(\mu_{S}(\mathbf{x, u}))$, $\forall \mathbf{x}\in\mathbb{X}, \mathbf{u}\in\mathbb{U}$ are the upper and lower bounds of the model imprecision respectively.
Although it is usually impractical to know $\mu_{S}(\mathbf{x, u})$ precisely, it is relatively easier to estimate its upper and lower bounds $\mu_{min}$ and $\mu_{max}$.

Due to the presence of unknown parameter $\mu$, $h: \mathbb{X}\rightarrow\mathbb{Z}$ from \eqref{eq:imprecise_function} is not a function, since a point in $\mathbb{X}$ now map to infinitely many points in $\mathbb{Z}$.
To find the likelihood function for the imprecise measurement model in \eqref{eq:range_meas2}, the measurement set $h(\mathbf{x, u};\mu)+w_{R}$ can be represented by a random closed set $\mathcal{Z}\subseteq\mathbb{Z}$ \parencite{10.5555/1512927}.
Then the generalized likelihood function characterized by the imprecise measurement function $h(\mathbf{x, u};\mu)$, which accounts for the model imprecision $\mu$ is defined as:

\begin{equation}
    \Tilde{L}_{R}(z; \mathbf{x, u}) =  
    Pr(z \in \mathcal{Z} ) = Pr(z \in  h(\mathbf{x, u};\mu) + \mathit{w}_{R} ) \label{eq:imprecise_likelihood1}
\end{equation}
where $Pr(\cdot)$ denotes the probability of an event.
In the following sections, we will refer to \eqref{eq:imprecise_likelihood1} as the imprecise likelihood function since $\mu$ represents the model imprecision.

When $\mathit{w}_{R}$ is zero mean white Gaussian, \eqref{eq:imprecise_likelihood1} can be solved analytically \parencite{5765431}:
\begin{equation}
    \Tilde{L}_{R}(z; \mathbf{x, u})  = \int^{\mu_{max}}_{\mu_{min}}\mathcal{N}(h; z, \sigma_{R}^{2})dh = \mathcal{C}(z; \underline{h}, \sigma_{R}^{2}) - \mathcal{C}(z; \overline{h}, \sigma_{R}^{2})
    \label{eq:RSSI_imp_likelihood}
\end{equation}
where $\mathcal{C}(z; \mu, \sigma^2) = \int_{-\infty}^{z}\mathcal{N}(\zeta; \mu, \sigma^2)d\zeta$ is the Gaussian cumulative distribution function (CDF), $\underline{h} = \underset{\mu}{\min}(h(\mathbf{x, u}; \mu))$ and $\overline{h} = \underset{\mu}{\max}(h(\mathbf{x, u};\mu))$.

We can understand the consequence of using an imprecise likelihood as having the effect of broadening the posterior PDF and imparting a higher variance compared to using the precise likelihood where the model imprecision $\mu$ is known~\parencite{5765431}. 
The imprecise likelihood enables us to expand the application scenarios of RSSI measurements for object tracking and localization  by simply providing upper and lower bounds of model imprecision $\mu$.

In addition to the RSSI measurement, we also consider obtaining 2-D (azimuth) angle of arrival measurements possible by planning a gyration motion by the drone as these are robust to variations in RSSI measurements impacted by complex terrain conditions. However, AoA measurements are still detected from the RSSI measurement receiver, and we propose a robust detection method and describe the AoA measurement model in the following section.

\subsubsection{Compensated AoA Detector and Measurement Model} \label{sec:aoa_model}
Since the UAV system we consider is highly maneuverable and is only equipped with a lightweight payload of a directional antenna and a simple receiver to provide RSSI measurements, we adopted the antenna rotation approach. 
The planning algorithm considers a measurement action involving the rotation of the antenna-equipped UAV to obtain the AoA of a radio signal. 
However, a problem arises under weak signals from distant radio tags. We describe the problem and our proposed compensation method to build a robust AoA detector and the associated measurement model below.

\noindent\textbf{Correlation coefficient based detector.~}The UAV performs one full rotation and uses the correlation between the collected RSSI measurements and the antenna gain pattern to determine the AoA. More specifically, after collecting $k$ detected RSSI measurements $\mathbf{z}_{R,t_1:t_k} = [z_{R,t_1},\dots,z_{R,t_k}]^{T}$ with associated detected object state $\mathbf{x}_{t_{1}:t_{k}} = [\mathbf{x}_{t_{1}}, \ldots, \mathbf{x}_{t_{k}}]$ and UAV state $\mathbf{u}_{t_{1}:t_{k}} = [\mathbf{u}_{t_{1}}, \ldots, \mathbf{u}_{t_{k}}]$ at time $[t_1,\dots,t_k]^{T}$,
the rotation AoA measurement is then given by:

\begin{equation}
    z_{A_{1}} = h_{A1}(\mathbf{x}_{t_{1}:t_{k}}, \mathbf{u}_{t_{1}:t_{k}}) = \underset{\alpha}{\operatorname{argmax}}~\rho(\mathbf{z}_{R,t_1:t_k}, \tilde{G}_{a}(\bm{\theta}_{t_1:t_k}^{(u)} + \alpha))  
    \label{eq:rel_bearing}
\end{equation}
where Pearson correlation coefficient $\rho(\mathbf{x}, \mathbf{y}) \triangleq \operatorname{cov}(\mathbf{x}, \mathbf{y})/(\sigma_{\mathbf{x}}\sigma_{\mathbf{y}})$, and $\bm{\theta}_{t_1:t_k}^{(u)} = [\theta^{(u)}_{t_1},\ldots,\theta^{(u)}_{t_k}]^{T}$ are UAV headings extracted from UAV states $\mathbf{u}_{t_{1}:t_{k}}$ (Recall that $\mathbf{u} = [\mathbf{u}_{l}, \theta^{(u)}]^{T}$).

While the correlation coefficient approach is sufficient when the receiver can detect the majority of radio pulses, its performance deteriorates as the strength of the detected signal reduces. This can significantly impact the ability to localize distant objects or objects equipped with radio tags configured with low transmit power. 
For a typical directional antenna gain pattern with two major lobes, its front (main) lobe pattern is generally similar to its back lobe. Due to the similarity in the front and the back lobe patterns, the measurement sequence for distant objects could correlate more strongly to the back lobe of antenna gain than the front lobe under a low number of detections (where not all of the measurements are detected due to the weak signal strength of received signals). Consequently, the correlation coefficient approach leads to an AoA measurement with approximately $\ang{180}$ error in these instances.

\begin{figure}[hbt!]
    \centering
    \includegraphics[width=0.5\textwidth]{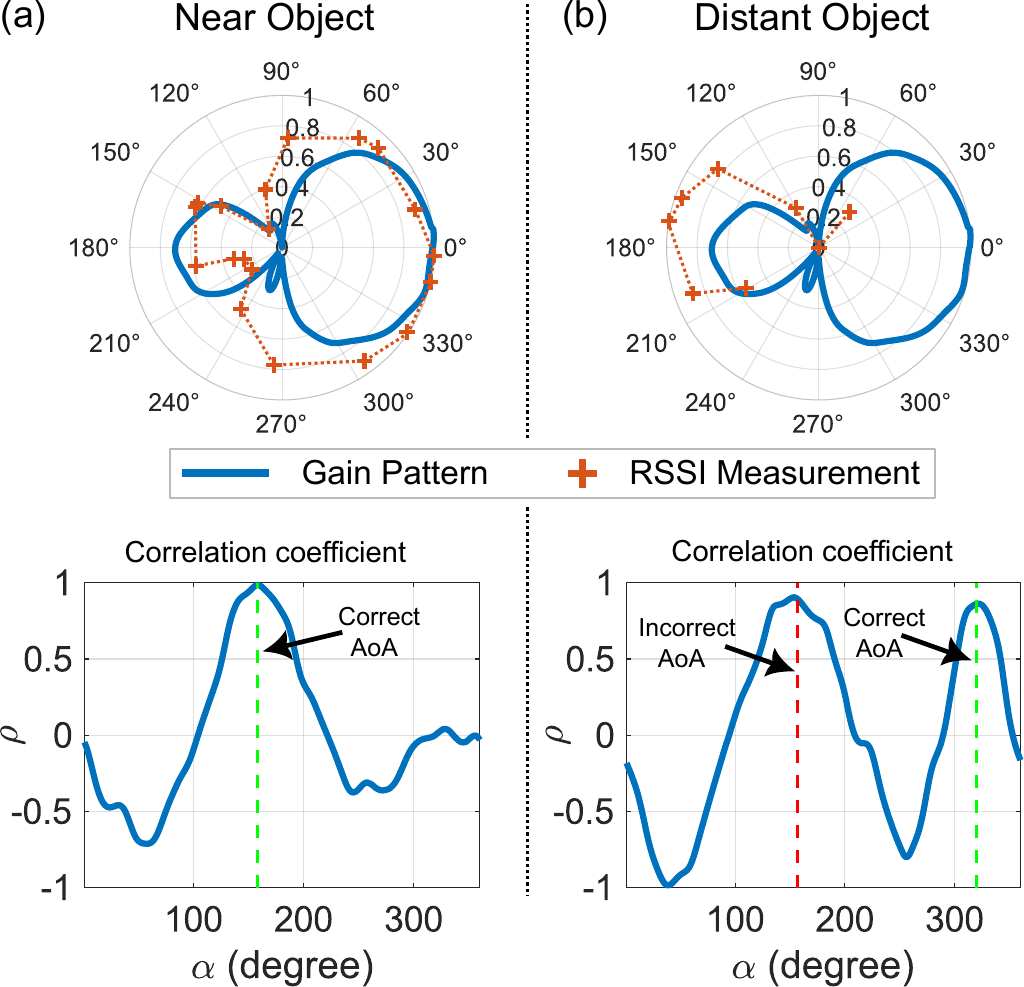}
    \caption{Illustrations of the correlation coefficient based AoA detector performance with RSSI measurements collected at different distances in field tests. (a)~For a close object, the majority of the signal emitted can be detected, and a correct AoA can be detected from the distinct peak in the correlation coefficient plot; (b)~For a distant object, not all signals emitted are detected. In this scenario, measurements correlate more strongly to the back lobe of antenna gain than the (main) front lobe. Consequently, an incorrect AoA is detected; we can observe an approximately $\ang{180}$ error.}
    \label{fig:corr_issue_illu}    
\end{figure}

Figure~\ref{fig:corr_issue_illu} illustrates a scenario from field testing the AoA measurement method where the receiver is only able to detect a fraction of the radio pulses emitted by the radio tag during a full rotation action when the distance between the UAV and the radio source is increased. In this scenario, the reduced number of measurements observed correlate more strongly to the back lobe of antenna gain than the (main) front lobe. Consequently, an incorrect AoA is detected. The result is an approximately $\ang{180}$ error.

\noindent\textbf{Cross correlation-based detector.~}In a scenario where the signal strength is weak, we can observe the receiver to be more likely to detect the signals when the main lobe (front) of the antenna is directed at the signal source. Therefore, instead of using the Pearson correlation coefficient, cross-correlation can be used to prioritize matching the strongest RSSI measurement to the front (main) lobe of the antenna:
\begin{equation}
    z_{A_2} = h_{A2}(\mathbf{x}_{t_{1}:t_{k}}, \mathbf{u}_{t_{1}:t_{k}}) = \underset{\alpha}{\operatorname{argmax}}~\langle\mathbf{z}_{R,t_1:t_k}, \tilde{G}_{a}(\bm{\theta}_{t_1:t_k}^{(u)} + \alpha)\rangle  \label{eq:cross_corr_bearing}
\end{equation}
where $\langle\cdot\rangle$ is the dot product, \textit{i.e.} $\langle a,b\rangle = \int a(x)b(x) dx$.

While \eqref{eq:cross_corr_bearing} can be used individually to generate AoA measurements with fewer outliers, from our observation in practice, when the radio tag's signal strength is strong, the cross-correlation method \eqref{eq:cross_corr_bearing} produces AoA measurements with higher variance than the correlation coefficient-based AoA detector. Hence, the sole use of a cross-correlation-based detector is detrimental to localizing radio tags approaching the UAV receiver. To overcome this issue, we introduce an approach to correct outlier AoA measurements.

\noindent\textbf{Compensated AoA detector.~}Based on the observations we discussed, we propose taking advantage of both AoA detection approaches to construct a more robust detector. We propose employing the cross-correlation \eqref{eq:cross_corr_bearing} method alongside the correlation coefficient method \eqref{eq:rel_bearing} to mitigate the AoA measurement ambiguity resulting from a correlation coefficient detector.

We propose exploiting the deviation between both AoA detectors to correct the correlation coefficient AoA detector measurements.
When an outlier AoA measurement is produced by the correlation coefficient AoA detector, the cross-correlation AoA detector's measurement will be significantly different. Notably, the correlation coefficient AoA detector can generate AoA measurement with $\ang{180}$ error for a typical directional antenna used for rotation-based AoA measurements such as Yagi antennas or the H-type antenna used in our experiments. Now, the compensated AoA measurement based on a decision threshold, $z_{A_{Th}}$, can be expressed as:
\begin{equation}
    z_{A} = \begin{cases}
    z_{A_1}& \text{if}~|z_{A_1} - z_{A_2}| < z_{A_{Th}} \\
    z_{A_1} - \pi~\SI{}{\radian} & \text{otherwise} \\
    \end{cases}
    \label{eq:combine_correlator}
\end{equation}

While we present the formulation of the method here, further discussion of experimental results from field tests to demonstrate the effectiveness of the proposed approach is presented in Section \ref{sec:exp_algorithm_validation}. Importantly, while the compensation method addresses the $\ang{180}$ ambiguity under noisy signal detection settings, if the number of measurements received is significantly low, then all of the AoA detection methods will fail to provide an accurate, usable measurement. We model the impact of such AoA measurement errors as \textit{noise} in our AoA measurement model.

\noindent\textbf{AoA measurement model.~}The distribution of $z_{A}$ is complex but can be approximated by a Gaussian distribution in practice \parencite{Cliff2018, UAV_RT} while assuming the object remains stationary during the rotation. Then:
\begin{equation}
    z_{A}\approx h_{A}(\mathbf{x}_{t_{k}}, \mathbf{u}_{t_{k}}) + \mathit{w}_{A}
\end{equation}
where$\mathit{w}_{A}\sim \mathcal{N}(\cdot; 0,\sigma_A^2)$ is a zero mean Gaussian noise with variance $\sigma_{A}^{2}$ and
\begin{equation}
    h_{A}(\mathbf{x}, \mathbf{u}) = \operatorname{arctan2}(\frac{u_{x}-p_{x}}{u_{y}-p_{y}})
    \label{eq:AoA_meas}
\end{equation}

Then the AoA likelihood is:
\begin{equation}
    L_{A}(z_{A}; \mathbf{x, u}) = \mathcal{N}(z_{A}; h_{A}(\mathbf{x, u}),  \sigma_{A}^{2}).
    \label{eq:AoA_likelihood}
\end{equation}

\subsection{Joint Measurement and Trajectory Planning Method}
In our proposed approach, the UAV not only needs to determine the trajectory action to navigate in the search environment but the measurement method to reduce the position uncertainty of wildlife to achieve a robust and rapid method of tracking and locating wildlife. 
The problem of automatically determining the best action can be formulated and solved efficiently under a POMDP framework.

Formally, a POMDP is defined by tuple $(\mathbb{X}, \mathbb{A}, \mathbb{Z}, \phi, \mathcal{R}, \mathcal{L})$. $\mathbb{X}, \mathbb{A}, \mathbb{Z}$ are state space, action space, and observation space respectively. 
$q(\mathbf{x'|x}, \mathbf{a})$ is the state transition function given action $a$ and current state $\mathbf{x}$, $\mathcal{R}$ is the reward function which characterizes the objective of the planner,  and $\mathcal{L}(\mathbf{Z|x}, \mathbf{a})$ is the observation likelihood function, where $\mathbf{Z}\in \mathbb{Z}, \mathbf{x}\in\mathbb{X}, a\in\mathbb{A}$. Considering the resource limitations and the need for an online planner for the tracking task, we consider a  computationally tractable POMDP formulation. Consequently, we employ a myopic planning strategy where the goal is to determine an optimal control action using a discrete action space at each planning iteration. Under a myopic planning strategy, the computational complexity is reduced by selecting one control action at a single planning iteration as opposed to considering multiple control actions in the future at multiple planning iterations. The optimal control action for a myopic planner $\mathbf{a}^{*}_{t}$ is defined by maximizing the expected reward function $\mathcal{R}_{t+H}(\cdot)$ over the action space:

\begin{equation}
    \mathbf{a}^{*}_{t} = \underset{\mathbf{a}\in\mathbb{A}}{\operatorname{argmax}}~ \mathbb{E}\left[\mathcal{R}_{t+H}(\mathbf{a})\right]
    \label{eq:total_reward}
\end{equation}

The following subsections describe essential elements of our POMDP-based planner and considerations that enable real-time planning decisions in the context of a UAV with limited onboard computing resources.

\subsubsection{Information-based Rewards} \label{sec:reward_func}
In a POMDP framework, the reward function can be categorized as i)~task-based rewards \parencite{task_reward1}; ii)~and information-based rewards \parencite{info_reward1}. A task-based reward is only applicable when the objective can be explicitly formulated. For object localization scenarios, the information-based reward is preferable to a task-based reward since the primary objective is reducing the position uncertainty of objects of interest, and information-based rewards prioritize gathering information; hence, has a strong relationship with the objective of improving the localization accuracy of objects~\parencite{Info_reward,Void_7967827}.
Importantly, in our problem formulation, information-based rewards provide a means to evaluate the quality of measurement type to aid the planner in deciding between taking RSSI or AoA measurements. Consequently, we considered three information-based reward formulations since a theoretical basis for determining the most effective formulation for our planning for tracking problems does not exist.

Given the predicted belief density of an object $\mathbf{\Psi}_{t+H|t} = \Psi_{t+H}(\cdot|\mathbf{Z}_{1:t})$ and the future updated posterior belief density $\mathbf{\Psi}_{t+H|t+H} = \Psi_{t+H}(\cdot|\mathbf{Z}_{1:t}, \mathbf{Z}_{t+1:t+H}(\mathbf{a}))$ at time $t+H$, where $\mathbf{Z}_{t+1:t+H}(\mathbf{a})$ is the hypothesized measurement set if action $\mathbf{a}$ was executed,
The three information-based rewards we considered are defined:

\begin{enumerate}
    \item R\'enyi Divergence \parencite{renyi1961} \parencite{RISTIC20101812}
    
        \begin{equation}
            \mathcal{R}_{t+H}^{(\text{R\'enyi})}(\mathbf{a}) = \frac{1}{\alpha - 1} \log \int \mathbf{\Psi}_{t+H|t}^{\alpha}\cdot \mathbf{\Psi}_{t+H|t+H}^{1 - \alpha} \delta\mathbf{X},
        \end{equation}
        where $\alpha \geq 0$ parameter determines the effect of the tails of two distributions on the rewards.

    \item Shannon Entropy \parencite{shannon}
    
        \begin{equation}
            \mathcal{R}_{t+H}^{(\text{Shannon})}(\mathbf{a}) = \mathcal{H}(\mathbf{\Psi}_{t+H|t}) - \mathcal{H}(\mathbf{\Psi}_{t+H|t+H}),
        \end{equation}
        where $\mathcal{H}(\Psi(\mathbf{X})) = -\int \Psi(\mathbf{X})\log{\Psi(\mathbf{X})} \delta\mathbf{X}$.
        
    \item Cauchy-Schwarz Divergence \parencite{cauchy}
    
        \begin{equation}
            \mathcal{R}_{t+H}^{(\text{CS})}(\mathbf{a}) = -\log \left[ \frac{\langle\mathbf{\Psi}_{t+H|t}, \mathbf{\Psi}_{t+H|t+H}\rangle}{\langle \mathbf{\Psi}_{t+H|t}, \mathbf{\Psi}_{t+H|t} \rangle \langle \mathbf{\Psi}_{t+H|t+H}, \mathbf{\Psi}_{t+H|t+H} \rangle} \right],
        \end{equation}
\end{enumerate}

\subsubsection{Measurement and Trajectory Planning Control Actions}
\begin{figure}[hbt!]
    \centering
    \includegraphics[width=0.4\textwidth]{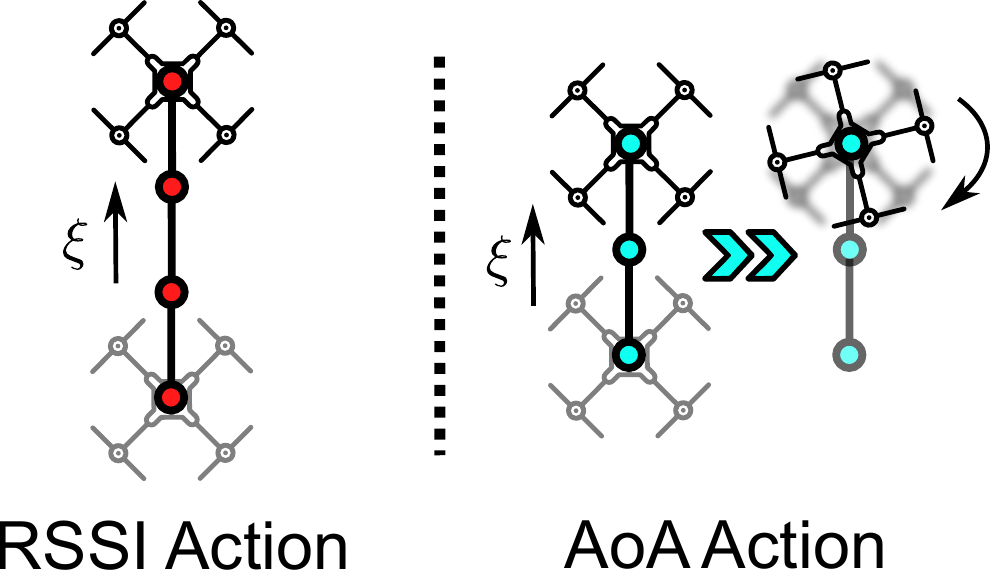}
    \caption{Illustration of an RSSI action in which the UAV travels along heading $\xi$ and an AoA action in which the UAV travels along heading $\xi$ followed by performing a full rotation maneuver.}
    \label{fig:action_illu}    
\end{figure}

One of the major strengths of employing the quad-copter UAV is the high maneuverability offered for traversing in a 3D space. Consequently, we have significant flexibility in designing the control action space. In our task, the available control actions should allow the UAV to explore the search area and collect RSSI or AoA measurements. To measure RSSI, the UAV does not require any special maneuvers, thus allowing for a change in heading to change the navigation path is sufficient. But, to measure AoA, a full rotation action must be executed by the UAV. Therefore, our action space for the planner can be decomposed into two sub-spaces, illustrated in Figure.~\ref{fig:action_illu}, and described below:

\begin{equation}
    \mathbb{A} = \mathbb{A}_{\text{RSSI}} \cup \mathbb{A}_{\text{AoA}}
\end{equation}

Recall that we consider a discrete action space to reduce the computational demands on the planner. 
Then, for the RSSI action space $\mathbb{A}_{\text{RSSI}}$ we define $n_{\xi}$ discrete headings, with each heading $\xi$ uniformly distributed across $\ang{0}$ to $\ang{360}$ (with reference to the geographical north), at a fixed altitude for $T_{P}$ seconds.

For the AoA action space $\mathbb{A}_{\text{AoA}}$, we augment the RSSI actions to include a full rotation action. Therefore, the AoA actions include two modes: i)~traveling along $n_{\xi}$ discrete, uniformly distributed headings at a fixed altitude for $T_{R1}$ seconds followed by: ii)~a full rotation maneuver with duration $T_{R2}$ seconds. 

The action space we define intentionally limits the UAV to a constant altitude, with two benefits: i)~it extends the flight time of the UAV as changing flight altitude consumes a substantial amount of energy; ii)~it simplifies the planning procedure. Further, allowing a change in altitude necessitates an obstacle avoidance component to ensure the safe operation of the UAV, which increases computation demands. By limiting the control action to only allow a UAV to travel on a 2D plain, we are able to ensure the UAV will not collide with obstacles as long as its initial altitude is higher than the tallest obstacle in the search area.

Notably, when evaluating the best control action between an RSSI action and an AoA action, their reward $\mathcal{R}_{t+T_{P}}(\mathbf{a}\in\mathbb{A}_{\text{RSSI}})$, $\mathcal{R}_{t+T_{R1}+T_{R2}}(\mathbf{a}\in\mathbb{A}_{\text{AoA}})$ cannot be directly compared unless both rewards are evaluated at the same horizon, \textit{i.e.}, $t+T_{P} = t+T_{R1}+T_{R2}$.
Therefore, we constraint the action space such that $T_{p} = T_{R1} + T_{R2}$ so that RSSI actions and AoA actions take the same amount of time to execute. Notably, when the UAV performs an AoA action, the receiver is still capable of measuring RSSI during the first traversal phase of the action. Hence, to avoid loss of useful information, the UAV also uses these RSSI measurements in addition to the AoA measurement generated at the end of executing the action to update the densities of objects.

\subsubsection{Void Constrained Trajectory Planning}
To minimize the disturbance of UAV operations to wildlife, we incorporated a void constraint into our planner.
The void constraint provides a probabilistic approach to maintaining distance to an object without knowing the exact state of the object.

\begin{figure}[hbt!]
    \centering
    \includegraphics[width=0.5\textwidth]{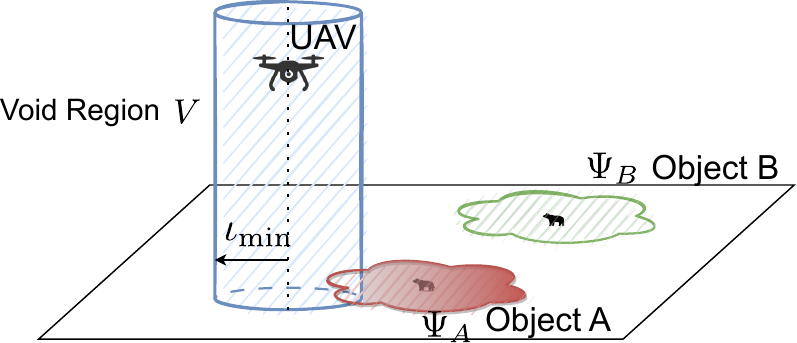}
    \caption{Illustration of a cylindrical void region $V$ with two object densities. The void constraint may be violated for object A if the probability of the non-overlapping region (red shaded area) between object A's belief density and void region is less than desired bound set by $P_{v\min}$.} 
    \label{fig:void_illu}    
\end{figure}

Given a region $S \subseteq \mathbb{X}$ and Bernoulli density $\Psi = (r, q(\cdot))$ on $\mathbb{X}$ where $p$ is approximated by set of weighted particles:
$q(\mathbf{x})\approx \sum^{N_s}_{i=1}\omega^{(i)}\delta(\mathbf{x} - \mathbf{x}^{(i)})$.
The void probability function can be expressed as \parencite{Void_7967827}:
\begin{equation}
B_{\Psi}(S) \approx (1-r) + r\cdot\big( 1 - \sum_{i = 1}^{N_s} \omega^{(i)}\mathbf{1}_{S}(\mathbf{x}^{(i)}) \big)
\label{eq:B_void}
\end{equation}
where $\mathbf{1}_{S}(\cdot)$ is the indicator function of the region $S$ equal to $1$ if $\mathbf{x}^{(i)} \in S$ and $0$ otherwise. We can interpret \eqref{eq:B_void} as the probability of an object with belief density $\Psi$ that is outside the region $S$. While the void region can be an arbitrary shape, we use a cylindrical void region with radius $\iota_{\min}$, where $V(\mathbf{u}, \iota_{\min})$ denote the void region given UAV state $\mathbf{u}$, given by:
\begin{equation}
    V(\mathbf{u}, \iota_{\min}) = \bigg\{\mathbf{x} \in \mathbb{X} \Big| 
                                         \sqrt{(p_{x} - u_{x})^{2} + (p_{y} -u_{y})^{2} } 
                                         < \iota_{\min} \bigg\},
\end{equation}
Now, the planning constraint can be expressed as:
\begin{equation}
	  \min\{ {B_{\Psi_{t}}} (V(\mathbf{u}_{t}, \iota_{\min} )) \} > P_{v\min}
	  \label{eq:void_constraint}
\end{equation}
where $P_{v\min} \in [0,1]$ is a user-defined void probability threshold. Figure~\ref{fig:void_illu} illustrates the resulting void region and the application of the constraint defined in \eqref{eq:void_constraint} with two object densities as examples.

Importantly, a cylindrical void region is a natural choice for two reasons: i)~when the radio signal is transmitted from a radio tag directly below the antenna, due to the orientation of the antenna onboard the UAV and the resulting lower gain, it can lead to increasing the missed detection rate. A cylindrical void region can potentially eliminate this scenario since a minimum horizontal distance can be maintained between the UAV and radio tags; and ii)~because the UAV will maintain a constant altitude during a mission to conserve limited onboard battery power. Provided the flight altitude is high enough, it is not necessary to consider the minimum vertical separation distance.

\subsubsection{Implementation Considerations for a Real-time System} \label{sec:planning_detail}
We considered a myopic planning formulation with a discrete action space to manage the complexity of the planning problem. To reduce the computational demands and realize a real-time planner without sacrificing the system's localization performance, we considered the two following approaches.

\emph{Planning for the closest unlocalized object.~} At every planning iteration, given a set of unlocalized objects' densities $\bm{\Psi} = \{ \Psi_{1}(\mathbf{X}), \ldots, \Psi_{n}(\mathbf{X}) \}$, instead solving for the optimal action that maximized the total reward for all densities $\bm{\Psi}$, we consider maximized the expected reward for the closest belief density $\Psi_{c}(\mathbf{X})$ to the UAV as adopted in \parencite{hoa2019jofr}.
where
\begin{equation}
    \Psi_{c}(\mathbf{X}) = \underset{\Psi(\mathbf{X})\in\bm{\Psi}}{\operatorname{argmin}}~ d(\bar{\mathbf{x}}, \mathbf{u})
    \label{eq:closest_density}
\end{equation}
with $\bar{\mathbf{x}}$ being the mean of object state given PDF $\Psi(\mathbf{X})$.
Once an object meets the condition to be considered localized, the object's belief density will be excluded from the next planning iteration and, therefore, reduce the number of densities the planner needs to process over time.

Planning for the closest unlocalized object has the benefit of reducing the computational complexity of calculating the reward. 
Limiting the number of planning densities makes the planner focus on computing actions that best minimize localization uncertainty for the closest object and, consequently, reduces the number of densities that need to be considered by the planner from $n$ to one. Interestingly, conservation biologists in the field also follow an identical strategy to locate multiple animals; they employ a handheld receiver system to home in on the closest perceived wildlife based on their determination of signal strength from audible \textit{beeps} from the receiver.

\emph{Predicted ideal measurement set (PIMS).~}In general, Monte Carlo integration is used to evaluate the expected reward in \eqref{eq:total_reward}.
This process requires drawing $M$ measurements $\mathbf{Z}^{(i)}_{t+1:t+H}(\mathbf{a}), i=1,\ldots, M$ which is obtained by sampling the belief density $\mathbf{\Psi}_{t+H|t}$ followed by generating a simulated measurement according to the measurement model.
Then the estimated expected reward is given by:
\begin{equation}
    \mathbb{E}[\mathcal{R}_{t+H}(\mathbf{a})] \approx \frac{1}{M}\sum_{i=1}^{M}\mathcal{R}^{(i)}_{t+H}(\mathbf{a})
    \label{eq:apporx_reward}
\end{equation}
As the number of samples $M$ increases, the estimated reward converges to the true expectation.
However, this method is computationally intensive, so instead, we adopted the \emph{predicted ideal measurement set} (PIMS) approach~\parencite{PIMS1} to compute the reward where only one instance ($M=1$) of future measurement set under an ideal condition is generated. The future measurement $\mathbf{Z}_{t+1:t+H}(\mathbf{a})$ is now computed by:
\textit{i)}~computing the expected state of the belief density $\mathbf{\Psi}_{t+H|t}$; and
\textit{ii)}~generating the expected measurement following the measurement function \eqref{eq:RSSI_meas} or \eqref{eq:AoA_meas} in the absence of measurement noise, false measurements, and miss detections ($P_D=1$).
Therefore, the estimated reward using PIMS is
\begin{equation}
    \mathbb{E}[\mathcal{R}_{t+H}(\mathbf{a})] \approx \mathcal{R}^{(\text{PIMS})}_{t+H}(\mathbf{a})
    \label{eq:pims_reward}
\end{equation}

Following the above implementation approach, the measurement and trajectory planning algorithm we developed is succinctly summarized  in Algorithm \ref{alg:planning}. 

\begin{algorithm}
\caption{Information theoretic \underline{me}asurement and \underline{t}r\underline{a}jectory planner (\ourMethod{})}  \label{alg:planning}
    \DontPrintSemicolon
    \SetKwInOut{Input}{input}
    \SetKwInOut{Output}{output}
    \SetKwInOut{Return}{return}
    
    \Input{Set of unlocalized objects' belief densities $\bm{\Psi}_{t}$, UAV state $\mathbf{u}_{t}$, action execution time $T_{p}$, reward function $\mathcal{R}(\cdot)$ , void radius $\iota_{\min}$, void probability threshold $P_{v\min}$, RSSI measurement action space $\mathbb{A}_{RSSI}$, AoA measurement action space $\mathbb{A}_{AoA}$}
    \Output{action $\mathbf{a}$}
  
    $\Psi_{t}(\mathbf{X}) \leftarrow \text{ClosestObject}(\bm{\Psi}_{t})$ \tcp*[r]{Find the closest belief density for the planning iteration, see~\eqref{eq:closest_density}}
    
    \ForEach{$\mathbf{a}^{(k)}\in \{\mathbb{A}_{RSSI}, \mathbb{A}_{AoA}\}$}{
        \For{$i = 1:T_{p}$}{
            $\bar{X}_{t+i} \leftarrow \mathbb{E}[\Psi_{t+i-1}(\mathbf{X})]$ \tcp*[r]{Compute the expected object state}
            $\mathbf{Z}_{t+i} \leftarrow h_{\mathbf{a}_{t}^{(k)}}(\bar{X}_{t+i}, \mathbf{u}_{t+i})$~\tcp*[r]{Generate simulated measurements using PIMS}
            $\Psi_{t+i}(\mathbf{X}) \leftarrow \text{BernoulliFiltering}(\Psi_{t+i-1}(\mathbf{X}), \mathbf{Z}_{t+i})$\tcp*[r]{Filtering, see ~\eqref{eq:bf_predict} and~\eqref{eq:bf_update}}
            
            \If{$\textsc{checkVoidConstraint}(\Psi_{t+i}(\mathbf{X}), \mathbf{u}_{t+i}, \iota_{\min}, P_{v\min})$}{
                $R^{(k)} \leftarrow 0$\;
                \textit{break}\tcp*[r]{Void Constraint violated. Evaluate next action}
            }
        }
        $R^{(k)} \leftarrow \mathcal{R}^{(PIMS)}_{t+T_{p}}(\mathbf{a}_{t}^{(k)})$\tcp*[l]{Compute the expected reward. See~\eqref{eq:pims_reward} where $H = T_p$}
    }
    $k \leftarrow \text{argmin}(R)$ \;
    \Return{$\mathbf{a}^{(k)}$}
\end{algorithm}

\section{Simulation Experiments} \label{sec:simulation_experiment}
To reduce development time and risks associated with evaluating concepts with a physical UAV system, and to investigate a wide range of settings we employed simulation-based experiments to evaluate our proposed approach to answer the following questions:
\begin{itemize}
    \item \textbf{Robustness under various terrain conditions.~}Our approach aimed to provide a robust, fast tracking and localization method under various terrain conditions. How does our proposed joint measurement and trajectory planning for tracking algorithm perform compared to existing approaches under different terrain conditions?
    \item \textbf{Impact of information-based reward functions.~}We investigated three different information-based reward function formulations. Does the choice of information-based reward functions provide a performance advantage?
    \item \textbf{Effectiveness and impact of void constrained trajectories.~}We employ a void constraint to maintain a safe distance between the UAV and our target wildlife. However, the void-constrained trajectories could impact the duration of a mission to localize wildlife. Is the approach effective and what is the impact on localization performance? 
    \item \textbf{Robustness under practical signal detection limitations.~}In the field, missed detections can occur and negatively impact performance. Hence, how does our approach perform under the different measurement detection probabilities compared to prior state estimation methods employed for localizing wildlife?
\end{itemize}

First, we elaborate on the complex VHF signal propagation model necessary to generate signals impacted by terrain conditions in our simulations, in Section \ref{sec:sim_model}.
Then, we describe the simulation settings in Section \ref{sec:sim_setup} and discuss the results from the simulations in Section \ref{sec:sim_results}.

\subsection{Complex VHF Signal Propagation Model} \label{sec:sim_model}
\begin{figure}[hbt!]
    \centering
    \includegraphics[width=0.5\textwidth]{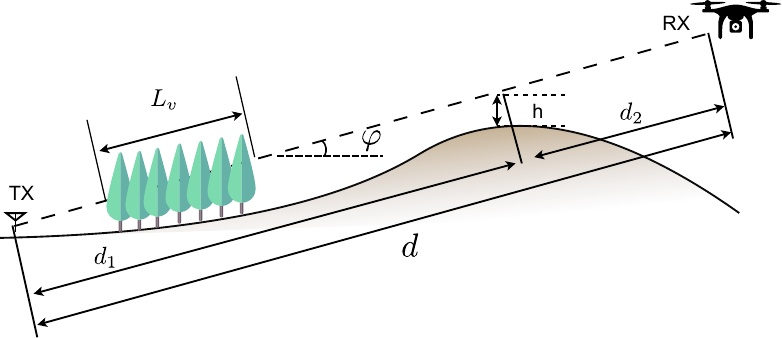}
    \caption{Illustration of simulated VHF signal propagation model and related parameters. Radio signal strength at receiver RX is influenced by the line-of-sight distance $d$, distance from the transmitter to most significant path blockage $d_{1}$, distance from the receiver to blockage $d_{2}$, height difference $h$ between the blockage and the path trajectory, vegetation depth $L_{v}$, and elevation angle $\varphi$}
    \label{fig:loss_illu}
\end{figure}

One of the key properties of our proposed formulation is the ability to relax the requirement for an accurate radio propagation model by incorporating an imprecise likelihood. To validate our approach in simulation settings, we employ a radio propagation model that captures the effects of: \textit{i)}~vegetation; and \textit{ii)}~terrain variations for generating the radio tag signals. We illustrate both types of signal propagation loss and parameters related to the model in Figure~\ref{fig:loss_illu} and briefly describe the formulation of the model used for generating signals below.

\paragraph{Vegetation Loss.}~We used the International Telecommunication Union (ITU) vegetation loss model to capture the effect of vegetation on VHF radio signals \parencite{ITU-Vegi}. The loss due to vegetation (assumed to be pine woodlands in our model) is defined as:
    \begin{equation}
        h_v = 0.25 f^{0.39} L_{v}^{0.25} \varphi^{0.05}
        \label{eq:vegetation_loss}
    \end{equation}
where $f$ is the signal frequency in $\SI{}{\mega\hertz}$, $L_{v}$ is the vegetation depth in meters, and $\varphi$ is the elevation angle in degrees. We use $f=\SI{150}{\mega\hertz}$ in our simulations. 

\paragraph{Terrain Shadowing and Diffraction.}~To model diffraction and shadowing effects from terrain conditions, we adopted the ITU terrain model where the additional loss is given by \parencite{ITU-Terrain}:
    \begin{equation}
        h_{d} = -20\cdot\frac{h}{F_{1}} + \SI{10}{\decibel}
        \label{eq:terrain_loss}
    \end{equation}
    where $h$ is the height difference between the most significant path blockage and the path trajectory, $F_{1}$ is the radius of the first Fresnel zone given by \parencite{ITU-Terrain}:
    \begin{equation}
        F_{1} = 17.3\cdot\sqrt{\frac{d_{1}d_{2}}{f\cdot d}} ~\SI{}{\meter},
    \end{equation}
    where $d$ is the distance between signal transmitter and receiver in $\SI{}{\kilo\meter}$, $d_{1}, d_{2}$ are the distances from transmitter and receiver to the blockage in $\SI{}{\kilo\meter}$ and $f$ is the signal frequency in $\SI{}{\giga\hertz}$. 
Overall, by subtracting \eqref{eq:vegetation_loss} and \eqref{eq:terrain_loss} from the ideal RSSI measurement model \eqref{eq:ideal_rssi_meas}, the propagation model that considers the complexities imposed by vegetation and terrain conditions related losses can be described as:
\begin{equation}
    h(\mathbf{x, u}) =  \tilde{\Gamma}(d_{0}) - \underbrace{10n\log_{10}(d(\mathbf{x, u})/d_{0}))}_{\text{distance loss}} + \tilde{G}_a(\zeta(\mathbf{x, u})) - \underbrace{h_v}_{\text{vegetation loss}} - \underbrace{h_{d}}_{\text{terrain loss}}
\end{equation}

An illustration of the impact of vegetation and terrain loss alone, over a physical terrain obtained from \parencite{DEM_australia} is presented in Figure.~\ref{fig:loss_illu_sim} in Section~\ref{sec:imprecise-rssi-model}.

\subsection{Comparison Approaches} \label{sec:sim_approaches}
We consider previous planning and measurement methods employed for tracking and locating wildlife to understand and evaluate the effectiveness of our proposed measurement and trajectory planning formulation for the problem.

We employed the Bernoulli filter formulation we proposed for the state estimator for all methods. This decision benefits all other methods because the filter formulation is inherently capable of dealing with practical issues such as miss detections. Importantly, employing the Bernoulli filter formulation for all comparison methods ensures the differences in performance are related to demonstrating that the proposed trajectory and measurement planning is a more effective approach for the task. This strategy can more clearly demonstrate the performance advantages gained from our proposed measurement and trajectory planning approach (abbreviated as \textit{\ourMethod{}}). Further, we use mobile radio tags to better capture wandering wildlife. Given the objective of void-constrained trajectories is to reduce disturbances, we employ void-constrained trajectory planning for \textit{all} comparison methods. In the following, we describe previous approaches and our specific improvements to facilitate a more useful comparison, especially in challenging settings.

\begin{itemize}
    \item \textbf{RSSI-only approach}. We adopt the RSSI-only approach described in \parencite{hoa2019jofr} where the UAV receives RSSI measurements from each radio tag, and the planner considers all future measurements RSSI only. The study in \parencite{hoa2019jofr} used a two-ray model to describe the propagation effect on RSSI over mostly flat terrains and was therefore not expected to perform well in more complex terrains. As a result, we \textit{introduced our imprecise RSSI measurement model} to improve the robustness of the original method in both filtering and planning algorithms. The RSSI-only approach with our proposed imprecise model is referred to as \textit{\rssiOnly{}} in the following sections.
    
    \item \textbf{AoA-only approach}. The AoA-only approach  uses rotation actions to acquire bearing measurements \parencite{rotate_aoa_nonrobot, monowing_aoa, VonEhr_SDR_AoA2016, Cliff2015, UAV_RT}. Instead of the AoA detector methods in prior work, we employed the \textit{improved} detector proposed in Section~\ref{sec:aoa_model} to generate AoA measurements. Further, for this approach, we propose using $\SI{20}{\second}$ to complete an AoA measurement, instead of the $\SI{45}{\second}$ described in prior work~\parencite{Cliff2015} using \textit{AoA only measurements}. Consequently, we keep the AoA measurement acquisition action and detector to that used in our \ourMethod{} in this setting; hence, we can expect the performance improvement to relate to our proposed measurement and trajectory planning approach in contrast to the detector improvements. We refer to this approach as \textit{\aoaOnly{}} to highlight the use of the proposed compensated AoA detector and the time duration for the measurement action.
    
    \item \textbf{AoA-with-RSSI-update approach}. The method described in \parencite{Cliff2018} sought to combine the benefits reported in \parencite{hoa2019jofr} with an AoA method. Here, rotation-correlation-based AoA measurements are used for object state estimation and trajectory planning, but an RSSI measurement update is also performed after generating each AoA measurement. The method in \parencite{Cliff2018} takes $\SI{45}{\second}$ to acquire a single AoA measurement and uses a log-path loss measurement model, given in~\ref{eq:RSSI_meas}, for the RSSI measurement update. Given the problems we have outlined in using RSSI models in complex terrains, we used a higher measurement noise in the log-path loss measurement model for hilly and mountain terrain to attempt to manage the RSSI model mismatch in complex terrains and ensure the comparison method remains competitive. We evaluate two variants.
    We employed implemented  \textit{\aoaRssiA{}}---using $\SI{20}{\second}$ for an AoA measurement---and  \textit{\aoaRssiB{}}---using $\SI{45}{\second}$ for an AoA measurement--to compare with potential advantages with a longer measurement duration selected in \parencite{Cliff2018}.
\end{itemize}

\subsection{Simulation Setup} \label{sec:sim_setup}
We describe the experimental settings and parameters employed in our extensive simulation-based study  below. Notably, we employed Digital Elevation Model (DEM) data from \parencite{DEM_australia} with $\SI{1}{\meter}$ resolution to create real-world terrain conditions for our experiments.
We consider three terrain conditions with increasing signal propagation complexity:
\begin{itemize}
    \item \textbf{Flat terrain.~}The data is obtained from Parkes, New South Wales (NSW), where elevation changes from $\SI{233}{\meter}$ to $\SI{239}{\meter}$. This terrain is representative of a simple environment for localization since the area has small elevation variations, and an accurate measurement model can be easily obtained for  state estimation. 
    \item \textbf{Hilly terrain.~}The hilly terrain is in Flinders Chase National Park, South Australia (SA), where the elevation changes from $\SI{40}{\meter}$ to $\SI{77}{\meter}$. The hilly terrain is more challenging than the flat terrain, given the higher elevation variation. 
    \item \textbf{Mountain terrain.~}The mountain terrain is in Rugby, NSW, where elevation changes from $\SI{595}{\meter}$ to $\SI{704}{\meter}$. The mountain terrain is the most challenging, given the large elevation variance and terrain obstructions.
\end{itemize}

\vspace{2mm}
\noindent\textbf{Settings.~}In each terrain, a UAV is tasked with localizing $N = 20$ mobile objects within a $\SI{2000}{\meter} \times \SI{2000}{\meter}$ area. Each radio tag object generates an RSSI measurement every $\SI{1}{\second}$. 
The initial state of UAV is $\mathbf{u}_{1} = [1, 1, 80+h_{0}, \pi/4]^{T}$ where $h_{0}$ is the elevation of terrain at the UAV's initial position. The UAV has a maximum velocity $v_{\text{max}} = \SI{10}{\meter/\second}$ and rotation angular velocity of $\pi/3~\SI{}{\radian/\second}$. A cylindrical void region with radius $\iota_{\min}=\SI{50}{\meter}$ and void probability threshold $P_{v\min}=0.95$ is used to constrain the UAV's trajectories.
For the RSSI measurement model, $\tilde{\Gamma}(d_{0}) = \SI{40}{\dBm}$, $n = 4$, $\sigma_{R} = \SI{4}{\decibel}$ are used. For the bearing measurement model, $\sigma_{A} = \SI{0.095}{\radian}$ is chosen. The rotation time to collect RSSI measurements to generate a bearing measurement is set to $\SI{20}{\second}$ except when evaluating the \aoaRssiB{} method in \parencite{Cliff2018}.
The measurement and trajectory planner evaluates actions every $T_{p} = \SI{30}{\second}$. For the R\'enyi divergence reward, $\alpha=0.1$ is selected based on the study in \parencite{hoa2019jofr}. 
The Bernoulli filter implementation, in all of the methods, uses birth probability $r_{b} = \num{1d-5}$, expected number of clutters $\lambda = 0.05$ with clutter density $c_{\text{RSSI}}(z) = \mathcal{U}[-120, 0]~\SI{}{\dBm}$ and $c_{\text{AoA}}(z) = \mathcal{U}[0, 2\pi]~\SI{}{\radian}$ for the clutter density of RSSI and AoA measurement updates, respectively---here, $\mathcal{U}[a,b]$ is continuous uniform distribution with interval $[a,b]$.

\vspace{2mm}
\noindent\textbf{Detection probability.~}To simulate the limited sensitivity of the radio receiver in practice, a sensitivity threshold $h_{Th} = \SI{-120}{\dBm}$ is implemented such that any simulated radio signal received with signal strength less than the threshold are discarded. 
Due to the effect of limited receiver sensitivity, the detection probability can vary as the state of UAV and radio tags changes.
Therefore, given the RSSI measurement model has Gaussian noise \eqref{eq:RSSI_meas}, we can express the detection probability $P_{D}(\cdot)$ of the radio signal as the following equation for use in the Bernoulli update step expressed in~\eqref{eq:bf_update}

\begin{equation}
    P_{D}(\textbf{x, u}) = \int_{h_{Th}}^{\infty}\mathcal{N}(z_{R};h(\mathbf{x, u}), \sigma^{2}_{R})dz_{R} = 1 - \mathcal{C}(h_{Th}; h(\mathbf{x, u}), \sigma^{2}_{R}),
    \label{eq:rssi_pd}
\end{equation}

here, recall that $\mathcal{C}(;)$ is the Gaussian CDF mentioned in \eqref{eq:RSSI_imp_likelihood}.

\vspace{2mm}
\noindent\textbf{Mobile radio tag objects.~}Radio tag objects to track and locate were randomly placed in the testing environment with elevation $\SI{0.2}{\meter}$ above ground. All objects were placed under vegetation with depth $L_{v} = \SI{1}{\meter}$ to generate complex VHF signal propagation artifacts and create challenging conditions for the proposed measurement and trajectory planner operating without an accurate measurement model and using an imprecise model instead. We modeled the object dynamics using a wandering model~\parencite{hoa2019tsp} with transitional density given by:
\begin{equation}
    q_{t|t-1}(\mathbf{x}_{t}|\mathbf{x}_{t-1}) = \mathcal{N}(\mathbf{x}_{t}; \mathbf{Fx}_{t-1}, \bm{\Sigma})
    \label{eq:object dymanic}
\end{equation}
where $\mathbf{F} = \mathbf{I}_{3}$ with $\mathbf{I}_{3}$ being $3\times 3$ identity matrix, $\bm{\Sigma} =  ~\text{diag}([2.5, 2.5, 0.0025]^{T})~\SI{}{\meter}^{2}$.

We assume the radio tag carried by each object emits an on-off-keying pulse signal with a unique frequency as illustrated in Figure~\ref{fig:raw_signal}. Therefore, each object can be uniquely identified by estimating its signal frequency and this significantly reduces computation-intensive data association procedures. 
Recall, to reduce the computational complexity of planning, the planning algorithm only selects optimal actions to reduce the estimation uncertainty associated with the object with the smallest Euclidean distance to the UAV. Once an object is considered localized, it will no longer be considered by the path-planning algorithm to control measurement and trajectory-planning actions. An object is considered localized when the estimation uncertainty has reduced to a sufficiently small level; for this, the determinant of its estimated covariance on the $x-y$ axis, $N_{\text{th}}$, being less than or equal to $\SI{2d4}{\meter\tothe{4}}$ was used in our simulations. 
We only use the $x-y$ axis to determine the termination condition for tracking because, generally, elevation resolution is not as important and to provide a fair comparison with the AoA method, which is not able to obtain elevation measurements.

\vspace{2mm}
\noindent\textbf{Performance measures.~}For each simulation, $100$ Monte-Carlo runs were performed. Given the key objectives of the UAV task, we used the following metrics to evaluate and compare the performance of each method.
\begin{itemize}
    \item \emph{Estimation Error.~}We are interested in the accuracy of estimating the location of wildlife. Hence, we use mean error of $N$ objects given by $\frac{1}{N}\sum_{i=1}^{N}\sqrt{(x_{truth}^{(i)} - x_{est}^{(i)})^{2} + (y_{truth}^{(i)}- y_{est}^{(i)})^{2}}$;
    \item\emph{Localization time.~}We want to minimize the flight time to locate radio-tagged wildlife since a UAV has limited onboard battery capacity and returning the UAV to the home base and changing batteries for a new mission is undesirable. Therefore, we measure the total time the UAV spends in the air to locate all objects.
\end{itemize}

\subsection{Simulation Experiments and Results}
\label{sec:sim_results}
\begin{figure}[hbt!]
    \centering
    \includegraphics[width=1.0\columnwidth]{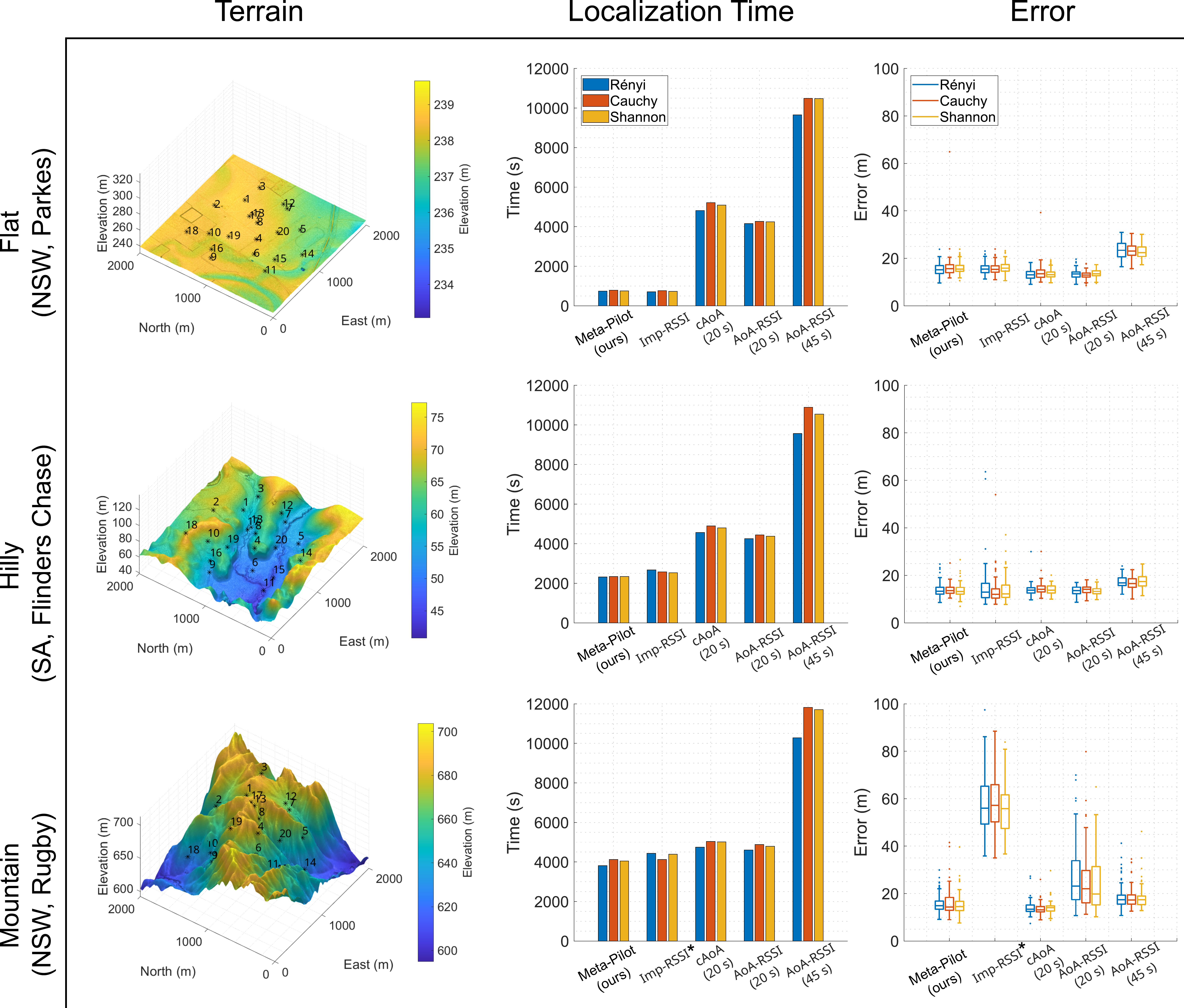}
    \caption{Comparing performance with mobile radio tags under various terrain conditions \textit{and} measurement and planning methods with different information-based reward functions under investigation. Performance is compared between different methods in terms of localization time and estimation error in flat, hilly, and mountain terrains over 100 MC runs where all of the planning methods use void-constrained trajectories. The time required for each AoA measurement is marked in brackets. (Note: \rssiOnly{}$^{*}$ method in mountain terrain uses a localization termination condition of $\SI{2d6}{\meter\tothe{4}}$) since the tighter uncertainty bound employed in other methods prevents localization of objects under less informative RSSI measurements.
    }
    \label{fig:all_compare}    
\end{figure}

\noindent\textbf{Robustness under various terrain conditions.~} These experiments aim to quantify advantages in terms of localization time and accuracy under different environments based on the information-based reward functions under investigation. 
We examined the performance of our proposed method in three distinct terrains: \textit{i)}~flat; \textit{ii)}~hilly, and \textit{iii)}~mountain terrains.

Figure~\ref{fig:all_compare} shows the Monte-Carlo simulation results in flat, hilly, and mountain terrains. Our proposed method, \ourMethod{}, is able to locate all $20$ radio tags consistently and faster---see localization times--than all AoA-based methods---\aoaOnly{}, \aoaRssiA{}, \aoaRssiB{}---across all terrain conditions  while maintaining low localization errors. Then, given the action choices the planner is able to make over the relatively flat terrain, where \textit{\rssiOnly{}} methods are expected to perform well, our proposed method was able to acquire the location of all $20$ radio tags as fast as the RSSI-based method whilst being significantly faster than AoA-based methods and without compromising localization accuracy. 

As the terrain complexity increases (Hilly to Mountain terrains), our method is able to select AoA measurement actions when necessary and achieve faster localization times than the Imp-RSSI and all AoA-based methods without sacrificing localization accuracy. In the mountain terrain, our planning for tracking approach performed better than all of the AoA-based methods---localization error is the same or better whilst the localization time is the least. Notably, in the mountain terrain, the \rssiOnly{} method is inherently not able to meet the tight termination condition set at $\SI{2d4}{\meter\tothe{4}}$ due to the complexity of the terrain. Therefore, a higher localization termination condition (more relaxed) of $\SI{2d6}{\meter\tothe{4}}$ is used for the Imp-RSSI method instead. As a result, we can observe the localization error of the method to be significantly larger than other methods.

\noindent\textbf{Impact of information-based reward functions.~}In the series of experimental results described in Figure~\ref{fig:all_compare}, all three reward functions resulted in approximately similar localization times and errors across all of the measurement and planning methods across all of the environments. However, R\'enyi divergence indicates a slight advantage in localization time. Hence, R\'enyi divergence was selected as the reward function to employ in field experiments.

\noindent\textbf{Robustness under practical signal detection limitations.~} In this experiment, we investigated the performance of our proposed approach under different measurement detection probabilities to understand the impact of our formation to address the practical problems experienced by signal detectors---missed detections. The parameters used in this experiment are those used for the hilly terrain and R\'enyi divergence was used as the reward function.
\begin{figure}[hbt!]
    \centering
    \includegraphics[width=0.4\textwidth]{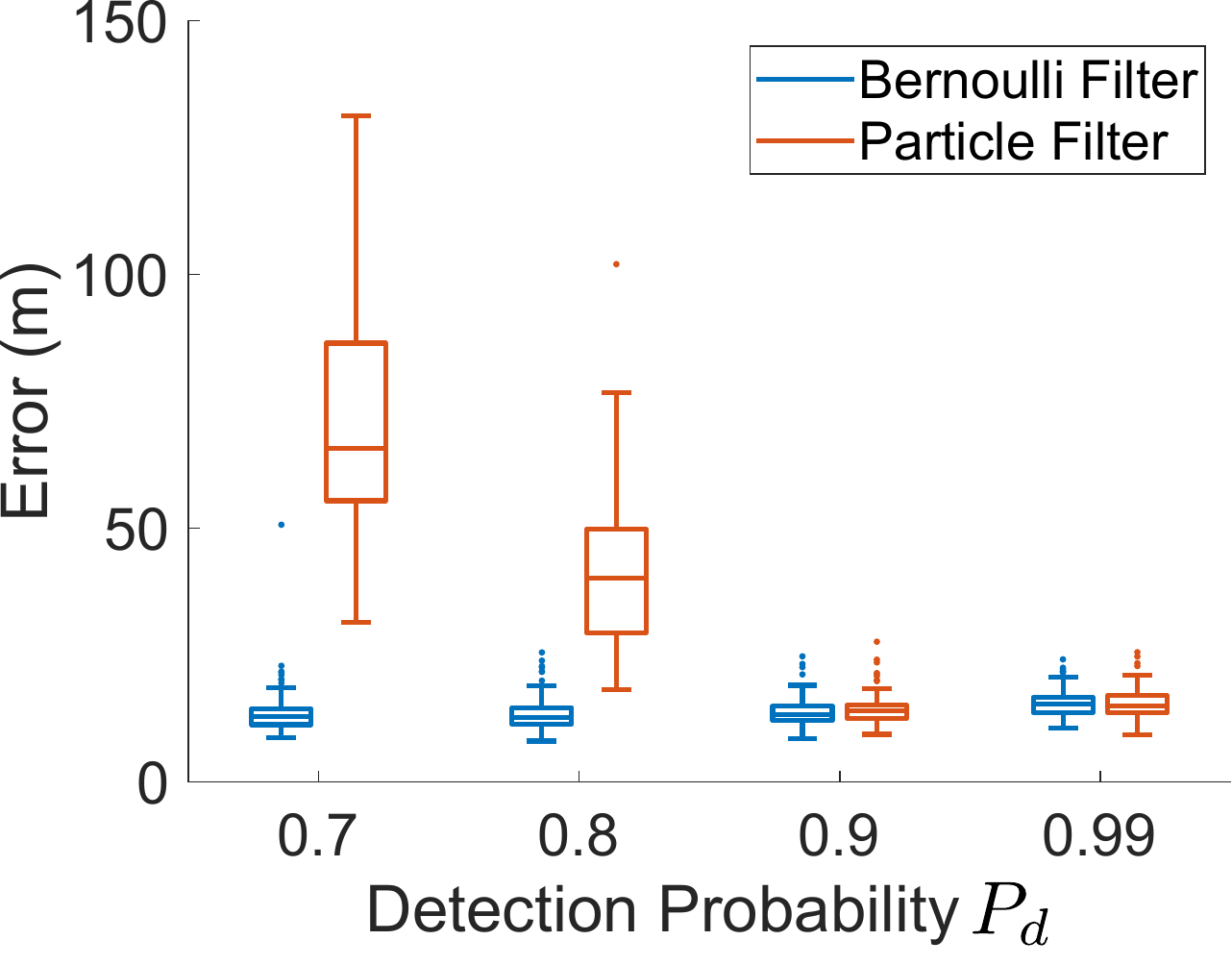}
    \caption{Detection   Probability   Impact   Experiments --- Comparison between Bernoulli filter (BF) formulation to explicitly consider miss detections and Particle filter (PF) under varying detection probability $P_D$ for the tracking and localization task over the hilly terrain with $100$ MC trials. The BF formulation is able to maintain a low localization error across different detection probabilities.
    }
    \label{fig:pd_comp}    
\end{figure}

Figure~\ref{fig:pd_comp} shows the Monte-Carlo-based comparison results of localization error between the Bernoulli filter and the particle filter (employed in prior work) implementations as the detection probability $P_D$ varies from $0.7$ to $0.99$. As $P_D$ decreases, the particle filter implementation suffers from low measurement detections, leading to high localization errors. 
In contrast, as expected, the Bernoulli filter formulation with explicit consideration for measurement detection probabilities is able to maintain a consistent localization accuracy under different detection probabilities.

\begin{figure}[hbt!]
    \centering
    \includegraphics[width=0.8\textwidth]{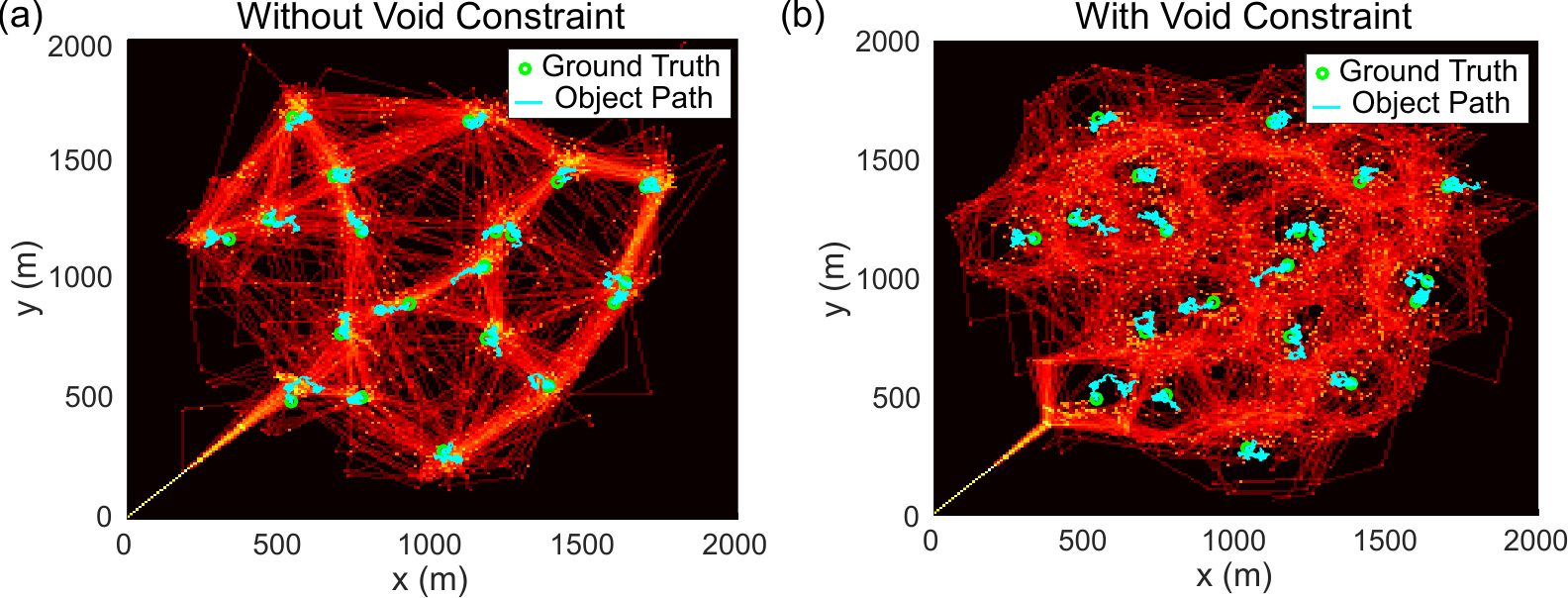}
    \caption{Void  Impact  Experiments With Mobile Radio Tags --- UAV trajectory heatmap over 100 MC trials. (a) Without void constraint; and b) with void constraint on the flat terrain. Green circles mark the initial ground truth location of mobile radio tags, while green paths denote the traversal paths of the radio tags.
    }
    \label{fig:void_heatmap}    
\end{figure}

\begin{figure}[hbt!]
    \centering
    \includegraphics[width=0.85\textwidth]{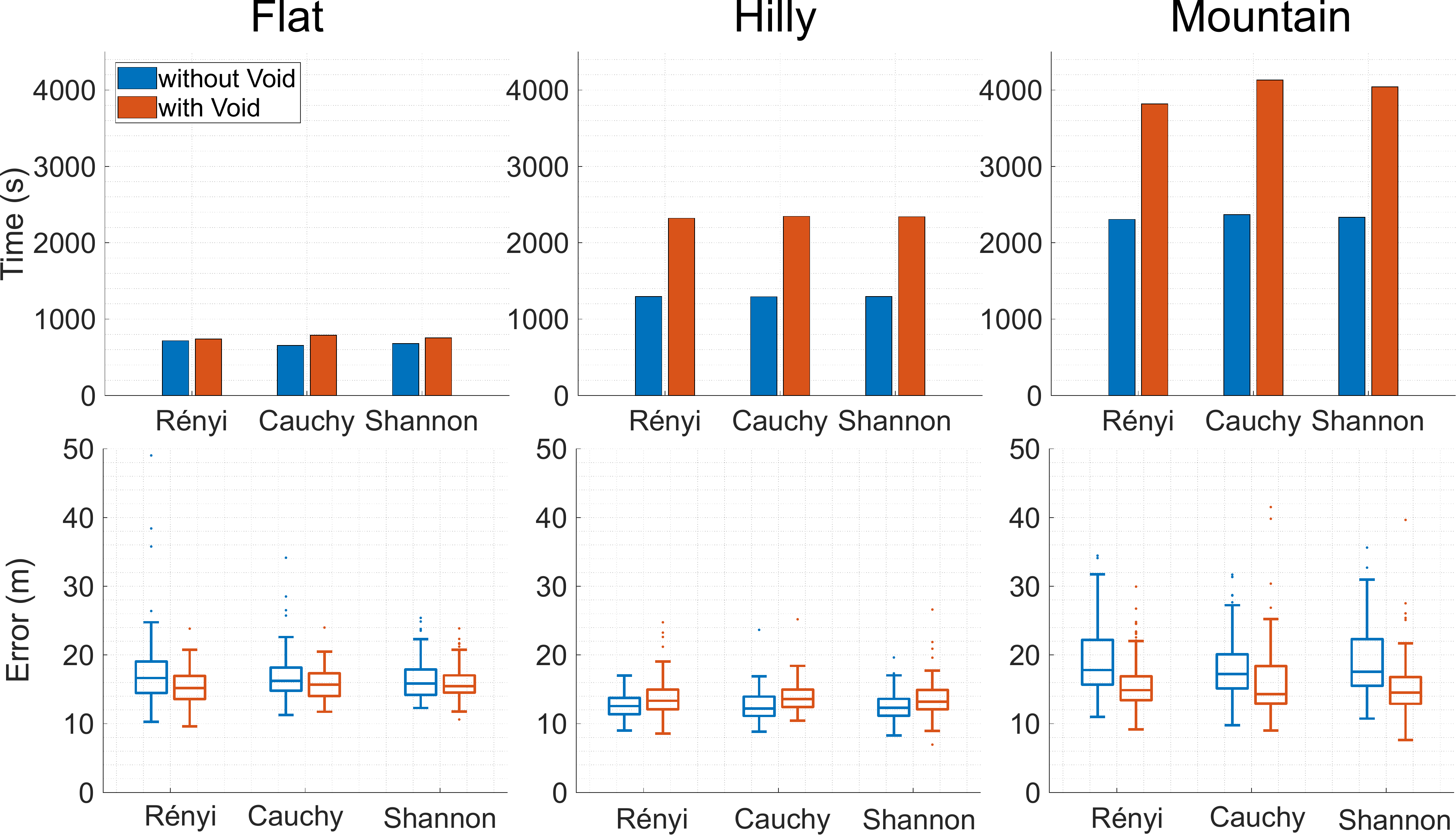}
    \caption{Void Impact Experiments --- Comparison with and without void constrained trajectories over $100$ MC trials in flat, hilly, and mountain terrain. }
    \label{fig:all_void}    
\end{figure}

\noindent\textbf{Effectiveness and impact of void constrained trajectories.~}The void probability functional constraint applied to planning aims to  minimize the disturbances to wildlife by distancing UAV trajectories away from the wildlife of interest. This experiment investigates the performance impact of void constraints in terms of localization time and error. 

Figure~\ref{fig:void_heatmap} illustrates $100$ trajectories for the UAV (generated over $100$ MC trials) in a task to localize $20$ mobile radio tags over the flat terrain. With the void constraint applied, as shown in Figure~\ref{fig:void_heatmap}(b), the flight path is changed noticeably  as the UAV attempts to maintain a safe distance to each radio tag. Since radio tags are mobile, only the initial ground truth position and their path are marked in the figure; hence, the mobility results in some trajectories appear to violate the void constraint. Notably, without the void constraint, the planner is expected to select the shortest path over the flat terrain, and the planner is expected to select RSSI-based measurement actions over AoA measurement actions, resulting in the UAV moving directly towards each radio tag.

Figure~\ref{fig:all_void} shows performance comparisons with and without void constraints on flat, hilly, and mountain terrains. In these terrains, with the void-constrained trajectories, the localization time with different reward functions all increase from $10 \%\sim 80 \%$ as the terrain becomes more complex. The increase in localization time is an expected result from both the void constraint trajectories and the less informative RSSI measurement model in the mountain terrain.
Due to the void constraint, the UAV needs to maintain a safe distance to radio tags and therefore requires spending extra time to navigate around each tag. In addition, in the mountain terrain, the RSSI measurement model is less effective in improving state estimation, while its use is conflicted by the need to maintain a minimum distance under void-constrained planning where the planner prevents the UAV from approaching the target to improve estimates. Consequently, the UAV is more reliant on time-consuming AoA measurements and this results in increased mission times for the localization tasks in more complex terrains under void-constrained planning. 
In contrast, over the flat terrain, the RSSI measurement imprecision is relatively low, and the measurements are more useful in improving state estimations even in the presence of trajectory constraints imposed by void preventing the approach of a UAV to a radio tag to obtain more informative RSSI measurements.

In terms of localization error, we can observe void-constrained trajectories to lead to comparable performance with  planning without void constraints. 
Interestingly, the results in the mountain terrain show that with void-constrained trajectory and measurement planning, the increased mission time to locate all radio tags has led to slight improvements in median localization accuracy.

\section{ConservationBot Prototype} \label{sec:prototype}
\begin{figure}[hbt!]
    \centering
    \includegraphics[width=1.0\columnwidth]{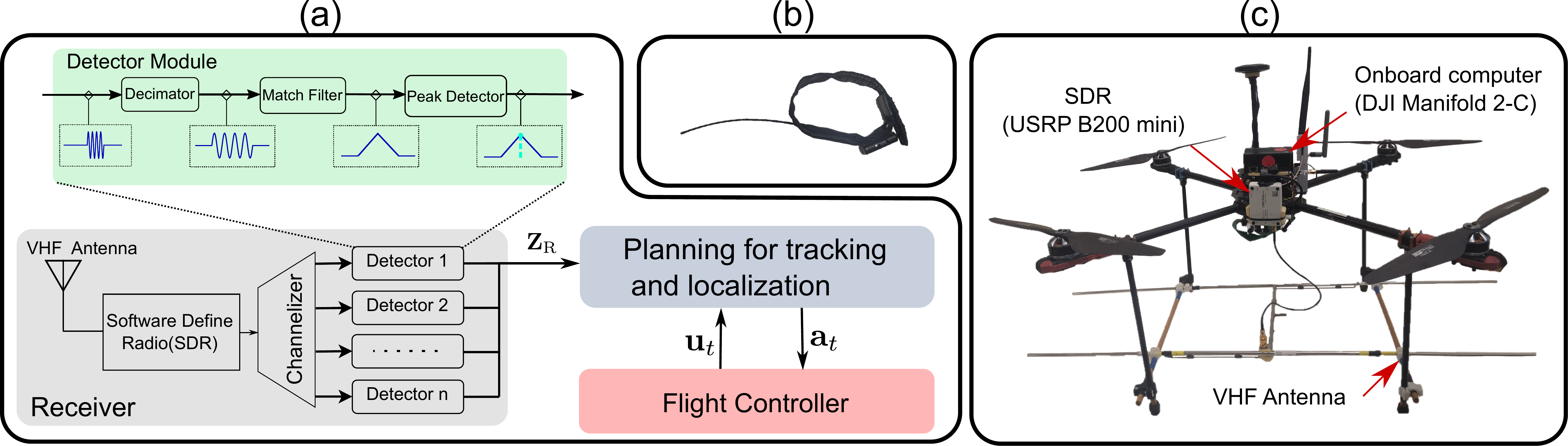}
    \caption{(a)~A system overview. The UAV state, control actions, and RSSI measurements are denoted by $\mathbf{u}_{t}$, $\mathbf{a}_{t}$ and $\mathbf{Z}_{\text{R}}$ respectively; (b) Lotek VHF Wildlife Radio Collar used in field experiments; (c)~ConservationBot.}
    \label{fig:system_block}    
\end{figure}

An overview of the prototype system---\textit{ConservationBot}--we built is shown in Figure \ref{fig:system_block}.
We employed a commercial directional H-type VHF antenna (Telonics RA-2AK) with $\SI{4}{\text{dBd}}$ gain and $\SI{10}{\decibel}$ front-to-back gain ratio and a Software Defined Radio (SDR) to construct the receiver. Given the significant advances in software-defined radios, their small form factor, and low weight (giving us the ability to reduce the payload of the sensor subsystem), we employed an SDR in our receiver. We selected the USRP B200mini-i because its lightweight has an adequately large $\SI{70}{\mega\hertz}-\SI{6000}{\mega\hertz}$ receiving frequency range, high sensitivity, and large bandwidth for simultaneously detecting multiple radio tags. We employed a DJI Manifold 2-C companion computer to implement the digital signal processing blocks of the receiver's detector module as well as the planning for tracking and localization algorithms.

The signal processing components of our receiver system are shown in Figure~\ref{fig:system_block}(a) and is implemented using GNURadio. The SDR is configured with a sample rate of $\SI{3}{\mega\sample\per\second}$.
We implemented matched filters to detect and measure the RSSI value of each radio tag.
The digitized RF data from the SDR are first channelized and down-sampled into a series of sub-channel with $\SI{80}{\kilo\hertz}$ bandwidth.
In each channel, the data is further decimated into $\SI{5}{\kilo\hertz}$ bandwidth signal to further improve SNR (signal-to-noise ratio) and reduce the computation complexity needed in later processing stages.
The data is then passed through a matched filter, and the RSSI value is identified by using a peak detector to generate the measurement $\mathbf{Z}_{\text{R}}$.

\section{Field Experiments} \label{sec:fieldexp}
We describe our extensive experimental regime to validate our approach and evaluate the performance of our aerial robot in the field. Our aims were to: 
\begin{itemize}
    \item Evaluate the detection range of the software-defined receiver architecture and hardware to understand the scanning range and the effectiveness of the proposed compensated AoA detector (Section \ref{sec:exp_system_cal} and \ref{sec:aoa_validation})
    
    \item Conduct field experiments to evaluate and compare performance between the proposed measurement and trajectory planning for tracking method with prior approaches (Section \ref{sec:exp_stationary} and \ref{sec:exp_mobile}) and illustrate the effectiveness of our approach (Section \ref{sec:exp_void_illustration}).
    
    \item Conduct field experiments to demonstrate the significant advantage provided by our aerial field robot over the manual methods employed for wildlife tracking (Section~\ref{sec:exp_human_compare}) and evaluate the proposed aerial field robot using Southern Hairy Nosed Wombats as a model species (Section \ref{sec:wombat_trial}).
\end{itemize}

We used Lotek VHF wildlife radio collars in our field experiments.
The radio collar is designed for continuous operation of $18$ months, and as a result of limited on-board battery power, its output power is limited to  $\SI{200}{\micro\watt}-\SI{500}{\micro\watt}$.
It transmits a $\SI{18}{\milli\second}$ pulse every $\SI{1}{\second}$ as illustrated in Fig. \ref{fig:raw_signal}.

\subsection{Software Defined Radio Receiver Detection Range} \label{sec:exp_system_cal}
\begin{figure}[hbt!]
    \centering
    \includegraphics[width=0.6\textwidth]{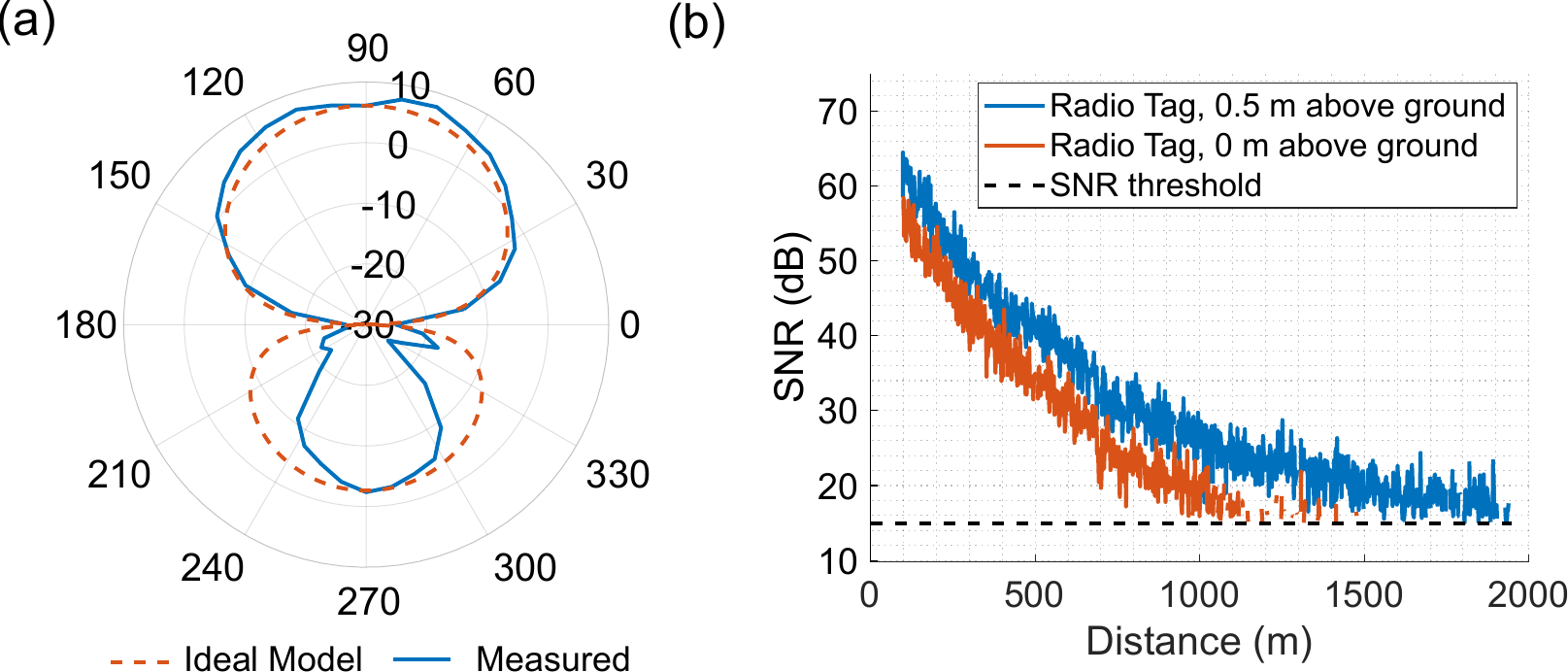}
    \caption{(a)~The measured gain pattern of antenna used by the receiver; (b) Detection range experiment: SNR at varying distances for radio collars placed at $\SI{0.5}{\meter}$ and $\SI{0}{\meter}$ above the ground measured at or above $\SI{15}{\decibel}$ SNR.}
    \label{fig:SNR_vs_distance}
\end{figure}
To understand the maximum detection range possible with our receiver architecture and hardware components, we performed multiple flights at a fixed $\SI{50}{\meter}$ altitude and measured the RSSI and SNR values of two radio collars placed at $\SI{0}{\meter}$ and $\SI{0.5}{\meter}$ above the ground. The $\SI{0.5}{\meter}$ height was chosen to represent the typically expected antenna height for above-ground wombats and the $\SI{0}{m}$ height represents a more challenging scenario where wombats  are at the entrance of their warrens and are also representative of smaller wildlife dwelling closer to ground level. During the flights, the heading of the UAV and antenna is fixed  and directed such that the antenna's maximum gain is directed towards radio collars. Importantly, the detection range demonstrates the scanning area possible for the receiver for a typically low-power, long-life, VHF radio collar---such as the one we used in our experiments---even without the ability of the platform to travel and cover a larger territory.

The detection distance is determined by the maximum distance between the receiver and radio collar when the received signal's SNR reached $\SI{15}{dB}$. Here, we used a conservative SNR level to yield minimal false alarms and a high detection probability. Consequently, in practice, a significantly longer detection range can be achieved and successfully employed given the capability of the Bernoulli filter formulation to accommodate false alarms and miss detections in real-world settings. 

Figure~\ref{fig:SNR_vs_distance}(a) shows the measured gain pattern $\tilde{G_{a}}(\cdot)$ of the antenna used in our system; the deviations from an ideal pattern in free-space is expected as the gain of the antenna is modified once mounted onto the UAV. Hence, the measured pattern is used in the measurement models we employ. Figure~\ref{fig:SNR_vs_distance}(b) shows the resulting detection range measurements. For the radio collar placed $\SI{0.5}{\meter}$ above ground, the signal can be reliably detected up to $\SI{2000}{\meter}$. As the height of the radio collar decreases, the system detection range reduced as expected, but even when the collar is placed directly on the ground, the range consistently exceeds $\SI{1000}{\meter}$.

\begin{figure}[hbt!]
    \centering
    \includegraphics[width=0.9\textwidth]{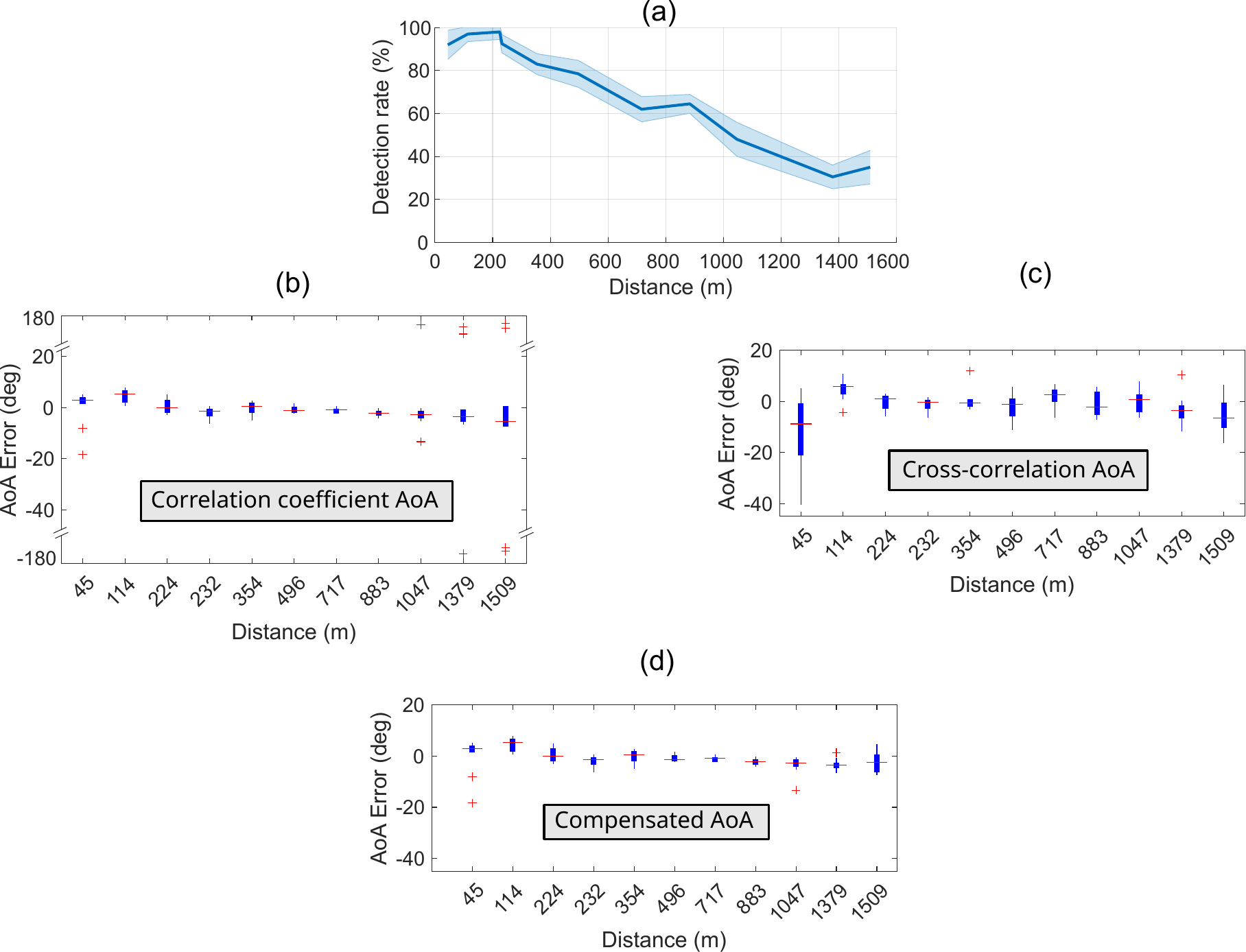}
    \caption{AoA measurements statistics at different distances. (a) Percentage of RSSI measurements collected during each AoA measurement, the shaded area shows one standard deviation. Each data point was built using AoA measurements of size $10$.  (b)~AoA measurement error using correlation coefficient \eqref{eq:rel_bearing}, error close to $\ang{180}$ can be observed when radio collar is above $\SI{1}{\kilo\meter}$. (c)~AoA measurement error using cross-correlation \eqref{eq:cross_corr_bearing}, no significant outlier is observed but overall has higher variance than (c) when radio collar distance is less than $\SI{1}{\kilo\meter}$. (d)~AoA measurement error using \eqref{eq:combine_correlator}.}
    \label{fig:AoA_error_stats}
\end{figure}

\subsection{Compensated AoA Detector Evaluation and Measurement Model Parameter Estimation} \label{sec:aoa_validation}
The AoA measurement errors can result from the accuracy of the UAV heading at the time of each RSSI detection since each AoA measurement requires the UAV to perform a full rotation and weak radio collar signals; for instance, at longer distances, we can expect the number of RSSI detections to reduce and potentially increase the AoA errors. 

To validate the effectiveness of our proposed compensated AoA measurement method and determine the AoA measurement noise variance, we collected 10 AoA measurements with a stationary radio collar at distances from $\SI{45}{\meter}$ to $\SI{1509}{\meter}$ while flying the UAV at a fixed altitude of $\SI{50}{\meter}$.
Figure~\ref{fig:AoA_error_stats} illustrates the AoA measurement errors and the percentage of detections.
As shown in Figure~\ref{fig:AoA_error_stats}(b), at distances larger than $\SI{1000}{\meter}$, the correlation coefficient based AoA measurement calculation produced outlier measurements with significant errors while the cross-correlation methods shown in Figure~\ref{fig:AoA_error_stats}(c) generates AoA measurements with relatively large variance. But the measurements are less sensitive to the detection rate.
The results in Figure~\ref{fig:AoA_error_stats}(d) demonstrate our proposed compensated AoA measurement method described in~\eqref{eq:combine_correlator}; we can observe the errors to be reasonably consistent across varying distances with small variances; notably, with the exception of one outlier at $\SI{45}{\meter}$, the majority of errors are within $\ang{10}$. Although the variation of AoA measurement noise can be modeled as a function of correlation coefficient~\parencite{Cliff2015} or distance, we opt for a fixed variance Gaussian noise model since the variation of AoA error with increasing distance is observed to be minimum with the compensated detector and detailed modeling would likely yield only marginal improvements.

\begin{figure}[hbt!]
    \centering
    \includegraphics[width=0.5\textwidth]{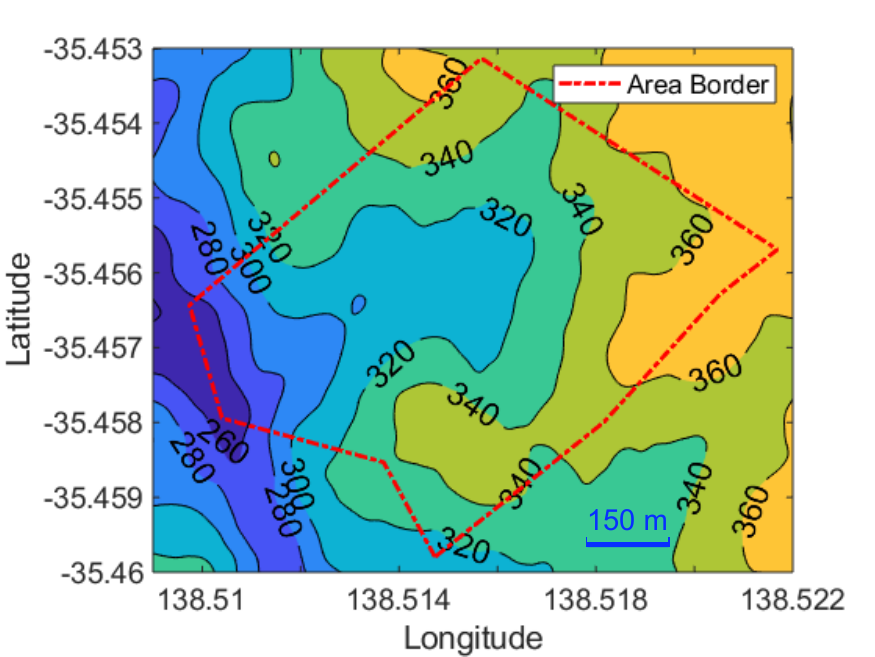}
    \caption{Contour map of hilly terrain field experiment site, Inman Valley, South Australia.}
    \label{fig:hilly_contour}
\end{figure}
\subsection{Field Experiments} \label{sec:exp_algorithm_validation}

 We conducted our field experiments to evaluate our prototype ConservationBot implementation over a more challenging, hilly terrain. The field experiments were conducted in the Inman Valley, approximately $\SI{10}{\kilo\meter}$ from Victor Harbor, SA, Australia---the area that the field experiments were conducted in was $\SI{40.86}{\hectare}$ in size. The terrain at this location is hilly and vegetated with remnant eucalypt forest to the height of approximately $\SI{12}{\meter}$ interspersed with thick shrubs to the height of approximately $\SI{1}{\meter}$, making it difficult for humans to traverse and an ideal environment to test our system.  Figure~\ref{fig:hilly_contour} shows the contour map of the field test area.

\subsubsection{Localization Performance} \label{sec:exp_stationary}
The first set of field trials was designed to evaluate the localization performance of our proposed \ourMethod{} method. For comparison, we selected \rssiOnly{} and \aoaOnly{} measurement methods \textit{without} measurement planning. Here, \rssiOnly{} method is able to account for uncertainty in the measurement model resulting from the hilly terrain and we can expect the approach to provide lower localization times due to the fast measurement acquisition times; and, the \aoaOnly{} with compensated AoA measurements minimize outliers and we expect the approach to provide improved localization performance because AoA measurement is less affected by terrain condition. (Please see Section~\ref{sec:sim_approaches} for detail descriptions of each method)

We selected a dispersed but stationary set of radio collars (placed at fixed locations).
This setting not only ensures the safety of the personnel involved but also allows us to design a consistent and repeatable experimental setting for conducting multiple missions to compare different approaches. A radio collar or radio-collared wildlife was considered localized if its location uncertainty, evaluated by the determinant of its estimated covariance $N_{th}$ is sufficiently small; we employed $N_{th}\le\SI{1d5}{\meter\tothe{4}}$ and imprecision range of $[-16, 9]~\SI{}{\decibel}$.
The ConservationBot was tasked to take off to $\SI{50}{\meter}$ above the launch position and execute the measurement and trajectory planning for the tracking algorithm to localize all radio-tagged objects.

Table.\ref{tab:hilly_result} summarizes time to complete missions and localization error results for robotic systems using only our proposed Imprecise RSSI model---Imp-RSSI---or AoA method with our proposed compensated AoA detector---cAoA(20~s)---in relation to our proposed \ourMethod{}. We found our proposed approach, \ourMethod{}, to provide the best set of consistent localization and total mission duration results (the set of missions with \ourMethod{} achieves the lowest mean error with  the smallest standard deviation and shortest mean localization time with the smallest standard deviation).

Unsurprisingly, as a rapid, yet simple, measurement acquisition method, the mean localization time of \rssiOnly{} is significantly better than the \aoaOnly{} method. Notably, we observed a similar result for \textit{comparable} terrains (the Flat and Hilly terrains) in our simulation study because the proposed imprecise RSSI model is able to account for the measurement model uncertainty and improve localization time and accuracy (see Fig.~\ref{fig:all_compare}). However, the ability to \textit{flexibly} employ more robust AoA measurement planning actions in \ourMethod{} to reduce object state estimation uncertainty achieved much better localization accuracy (highest mean and smallest standard deviation demonstrating consistent performance) and, on average, shorter, more consistent (smaller standard deviation) mission completion times compared to \rssiOnly{}.

In contrast, as expected, whilst also confirming our results in the simulation study for the Hilly terrain, \ourMethod{} significantly outperformed the \aoaOnly{} method in terms of localization time in field trials. \ourMethod{} required only one-third of the time on average to successfully locate the four stationary radio collars with better mean localization accuracy compared to the \aoaOnly{} method. AoA measurements are robust to impacts from the multiplicity of VHF signals over complex terrains but it is a costly action for the robot to perform in terms of both time consumed and energy expended over a \textit{in situ} rotation. But, our \ourMethod{}'s ability to dynamically decide when to action AoA measurements significantly minimized the cost implications of employing the more robust AoA measurements whilst still benefiting from the measurement method.  Consequently, \ourMethod{} achieves a significant reduction in mission completion times in contrast to the \aoaOnly{} method but without impacting localization accuracy.


\subsubsection{Tracking and Localizing Mobile Objects} \label{sec:exp_mobile}
In this set of field experiments, we employed the same settings used in Section~\ref{sec:exp_stationary}, with the exception of having two of the four VHF radio collars mobile to validate the capability of ConservationBots to track and locate mobile radio collars. Here, two VHF radio collars were carried by human volunteers tasked with performing a wandering motion from their starting locations at approximately $\SI{1}{\meter/\second} \sim \SI{2}{\meter/\second}$. The trajectory of the mobile objects was captured using a phone-based GPS data logger and later compared to the reported object location to obtain the localization error.

Table.~\ref{tab:hilly_result} summarizes the results for localizing mobile objects.
The results demonstrate that our proposed \ourMethod{} can track and locate mobile and stationary objects. When comparing mobile objects to the results for localizing only stationary objects, we observed a decrease in accuracy and a slightly larger variation in localization times as the moving radio collars can impact the time to reduce the uncertainty associated with estimated positions to a desirable level whilst also planning for void constrained trajectories.

\begin{table}[hbt!]
\caption{Comparison of localization performance for stationary and mobile objects in a \textit{hilly} environment where our ConservationBot was configured with: i)~our proposed imprecise RSSI model alone; ii)~compensated AoA detector alone as described in Section~\ref{sec:sim_approaches}; and iii)~our proposed measurement and trajectory planner with the imprecise RSSI model and compensated AoA detector.} 
\small
    \centering
    \begin{tabular}{lcccc}
        \toprule
         Method & Setting & Trials & Error $\pm~1\sigma$  (\SI{}{\meter}) & Time $\pm~1\sigma$  (\SI{}{\second}) \\
         \midrule
         \begin{tabular}{@{}l@{}}Measurement and Trajectory Planner \\ (\textbf{\ourMethod{}}) \end{tabular}
         & Stationary radio collars& 8 & $\mathbf{34.5 \pm 8.5}$ & $\mathbf{231 \pm 23}$ \\
         \begin{tabular}{@{}l@{}} Imprecise RSSI only \\ (\rssiOnly{}) \end{tabular}
         & Stationary radio collars& 8 & $43.3 \pm 11.2$ & $245 \pm 30$ \\
         \begin{tabular}{@{}l@{}} Compensated AoA only\\ (\aoaOnly{}) \end{tabular}
         & Stationary radio collars & 8 & $40.0 \pm 12.7$ & $745 \pm 123$ \\
        \midrule
        \begin{tabular}{@{}l@{}} Measurement and Trajectory Planner \\(\textbf{\ourMethod{}}) \end{tabular}
        & \begin{tabular}{@{}c@{}}Mobile \& Stationary \\ radio collars\end{tabular} & 8 & $45.1 \pm 17.5$ & $230 \pm 83$  \\
        \bottomrule
    \end{tabular}
    \label{tab:hilly_result}
\end{table}

\begin{figure}[hbt!]
    \centering
    \includegraphics[width=0.95\columnwidth]{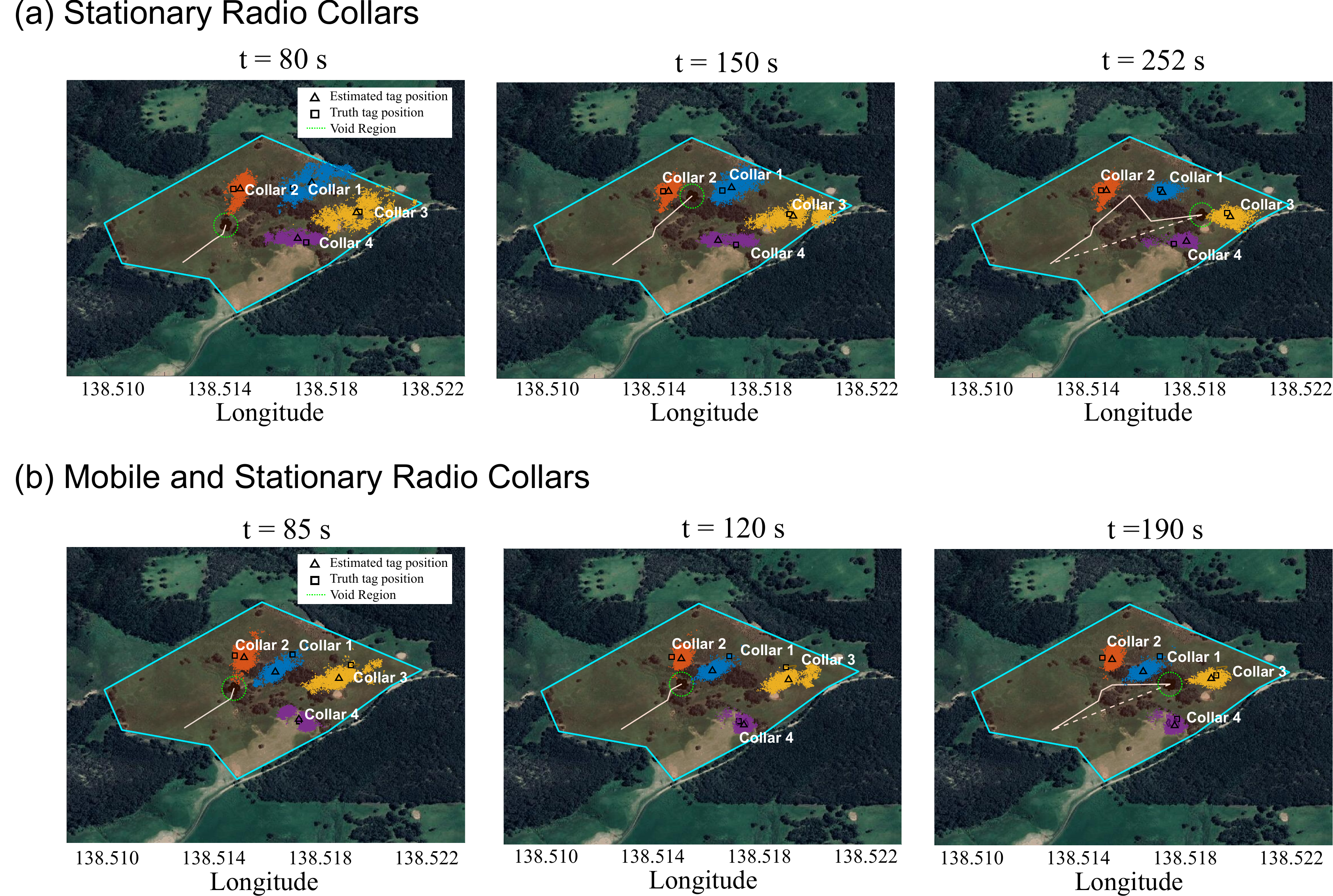}
    \caption{
    Instances of intermediate belief density representing the estimated location of radio collars for two scenarios selected from our field trials: (a) localizing $4$ stationary VHF radio collars, (b) localizing $2$ stationary collars (1 and 2) and $2$ mobile collars (3 and 4). The effect of void-constrained trajectories can be observed as the UAV navigates around (unlocalized) radio collars to maintain a safe distance. Here we can observe the convergence of belief densities of all radio collar location estimates and the operation of the planner generating trajectories to maintain the void constraint.}
    \label{fig:static_mobile_example}    
\end{figure}
\subsubsection{Minimizing Disturbances with Void Constrained Trajectories} \label{sec:exp_void_illustration}

We present two missions as examples to illustrate the progression of the tracking and localization task and the manner in which the void-constrained trajectories are able to maintain a safe distance from the VHF radio collars of interest. Figure~\ref{fig:static_mobile_example}(a) depicts the evolution of belief densities for each radio collar over time. From these snapshots, we can observe a typical trajectory and behavior as a result of the void constraints. 
At time $t = \SI{80}{\second}$, the UAV is focusing on locating collar $2$. Shortly after collar $2$ is located, the UAV proceeds to navigate toward the next closest collar (collar $1$). We can observe that during this process, the majority of collar $2$'s belief density (represented by orange particles) remains outside the void region of the UAV (green dashed circle). After time $t=\SI{150}{\second}$, the UAV finishes locating collar $1$ and subsequently heads toward collar $3$ while locating collar $4$ during the process and return to the home-base location after all collars have been found at time $t=\SI{252}{\second}$. Here the effect of void constraint becomes more prominent as illustrated by the UAV navigating around collar $1$.

Figure~\ref{fig:static_mobile_example}(b) shows another instance of intermediate belief densities for the task of locating two stationary (collar $1-2$) and two mobile (collar $3-4$) VHF radio collars. A similar planning strategy illustrated in Figure~\ref{fig:static_mobile_example}(a) can also be observed here.  At time $t=\SI{85}{\second}$ the UAV moves towards the collar determined to be the closest (collar $2$). However, due to the void constraint, the UAV is unable to move too close to collar $2$ and instead plans a trajectory around collar $2$ as seen at time $t = \SI{120}{\second}$. Then, after both collar $1$ and $2$ are located, the UAV traverses towards collar $3$ and completes locating all of the collars at time $t=\SI{190}{\second}$. These results from field experiments demonstrate that our proposed planning method is able to locate all stationary or mobile radio collars while maintaining a safe distance. 

\begin{figure}[hbt!]
    \centering
    \includegraphics[width=0.5\textwidth]{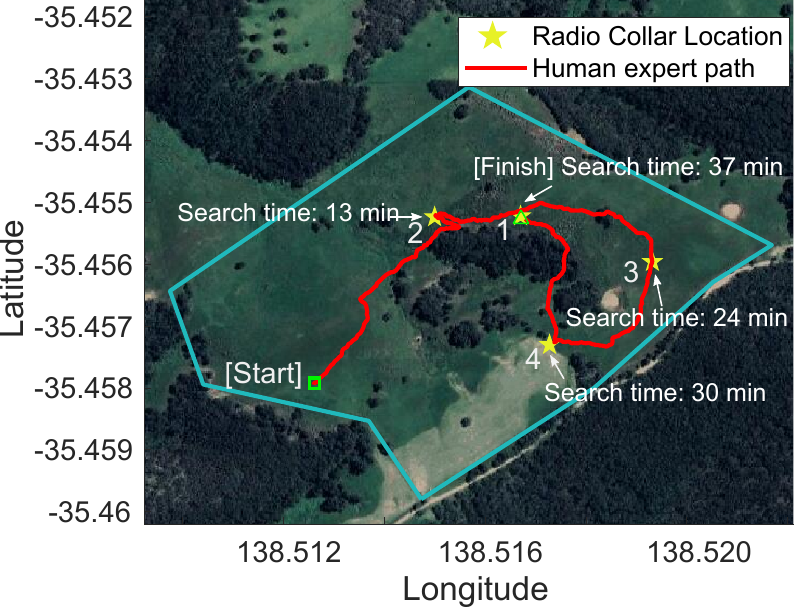}
    \caption{Results from conducting the tracking task with a human expert with over 20 years of field experience using the manual method with $4$ stationary VHF radio collars in a \textit{hilly} environment. Radio collars are found in the order of $2, 3, 4, 1$ by the expert human tracker. Completing the localization task took $37$ minutes for the human expert and only $4$ minutes for our ConservationBot.}
    \label{fig:manual_Result}
\end{figure}
\subsubsection{Comparisons with a Human Expert} \label{sec:exp_human_compare}
In order to demonstrate the benefits of the autonomous method of tracking wildlife we developed here using our Conservationbot, compared to traditional wildlife tracking methods (involving a field scientist trekking, often through difficult terrain, carrying bulky radio-telemetry equipment), we invited an expert conservation biologist with over 20 years wildlife tracking experience, to compete with our robotic platform.
To ensure comparable settings, $4$ stationary radio collars were used in this experiment. The human expert was given the list of radio collar frequencies at the start but had no prior knowledge of the positions of the radio collars. The human expert and the ConservationBot set off from the same starting position. 

The localization time for each radio collar and the traveled path of the human expert is shown in Figure~\ref{fig:manual_Result}. As expected, the results show a significant difference in search and localization time between the manual method and the ConservationBot, demonstrating the effectiveness of the ConservationBot as a field robot for the task. Human expert, first localized collar $2$ after $13$ minutes of search time (notably, during this time-lapse, the ConservationBot has located all 4 radio collars and returned to the home base).
Due to the terrain and vegetation coverage impacts on VHF signal propagation, collar $3$ was selected by the human expert as the next collar to localize, although collar $1$ was closer. Completing the localization task took $37$ minutes; this was significantly higher than the  $4$ minutes required for our ConservationBot. Notably, the equipment used by the human tracker was superior to that employed on the ConservationBot; specifically, a three-element Yagi antenna with a higher gain and front-to-back ratio, compared to the lower gain two-element model used by the ConservationBot, along with a more sensitive radio receiver---an Australis 26K radio receiver---was used by the human expert.

\subsection{Field Trials with Southern Hairy Nosed Wombats} 
\label{sec:wombat_trial}
\begin{figure}[hbt!]
    \centering
    \includegraphics[width=0.45\columnwidth]{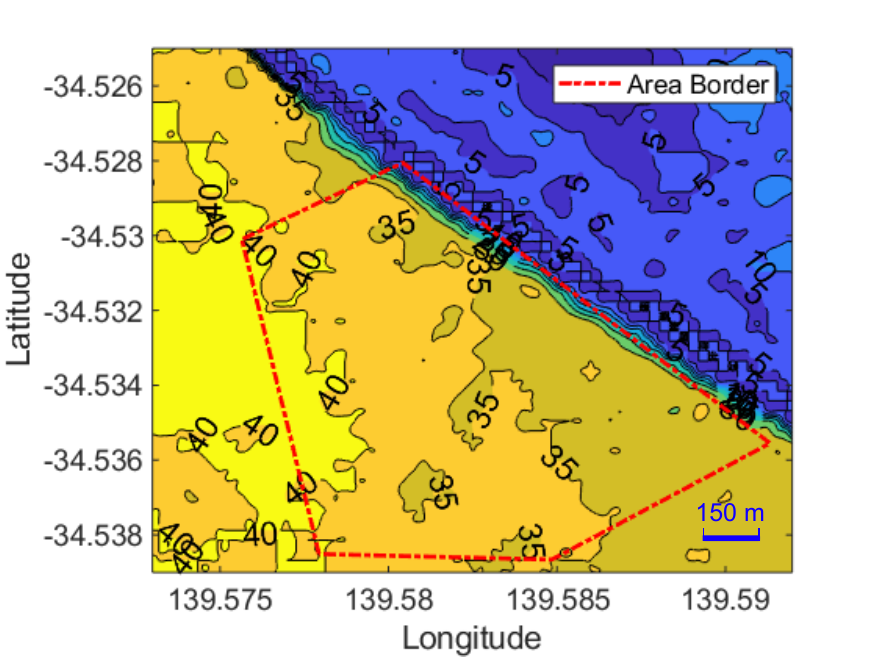}
    \captionof{figure}{Contour map of wombat habitat terrain in Swan Reach, South Australia.}
    \label{fig:wombat_contour}
\end{figure}

We participated in a field experiment where our ConservationBot was deployed to localize radio-tagged wombats in a conservation project. The trials were performed near Swan Reach SA, Australia. A total of $6$ southern hairy-nosed wombats (Lasiorhinus latifrons) were captured, radio-tagged, and released prior to the trials. The terrain at this location is comprised of remnant mallee vegetation interspersed with native grasslands, eucalyptus trees and is, thus, representative of flat terrain with less than $\SI{5}{\meter}$ elevation variations across the site as shown in Figure~\ref{fig:wombat_contour}.

Trials were conducted during the daytime to comply with university health and safety regulations, legal and risk implications, as well as regulations and procedures governing the testing of autonomous aerial vehicles. Southern hairy-nosed wombats, a nocturnal species, are usually less active and located in warrens underground \parencite{wombat_underground} during daylight hours.
While this behavior meant that wombats were mostly stationary, the radio signal would be greatly attenuated by the ground resulting in a significant reduction in the maximum detection range of signals. These attributes provided a very challenging setting for a field trial.
The UAV was launched to a fixed altitude of $\SI{50}{\meter}$ and tasked to localize all of the detectable radio-tagged wombats. Subsequently, manual tracking was undertaken to determine the ground truth of each wombat's location and compare it to the reported location of our system to determine the reported accuracy.

\begin{figure}[hbt!]
    \centering
    \includegraphics[width=1.0\columnwidth]{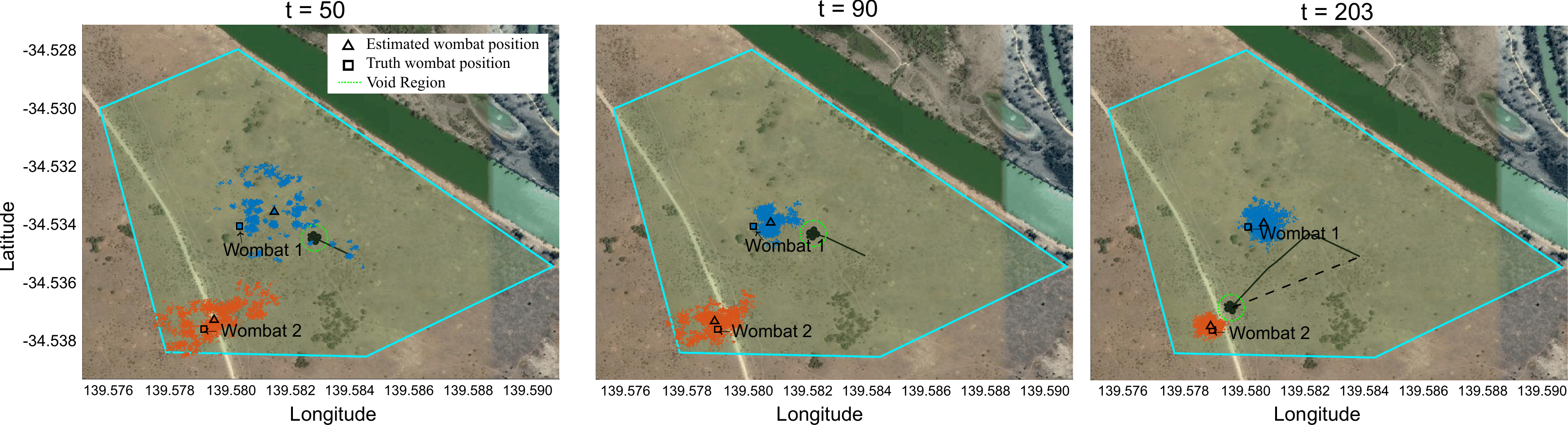}
    \caption{Intermediate distributions of belief density representing the estimated location of the underground VHF radio collared southern hairy-nosed wombats. The UAV first moves toward Wombat~1 ($t = 50$ to $t = 90$). Then moves around Wombat~1 due to the void-constrained trajectory planner and navigates to Wombat~2 after Wombat 1 is localized ($t = 203$). The $\square/\triangle$ denotes the truth---determined by manual tracking methods--and estimated wombat positions by the ConservationBot, respectively; the green dashed line denotes the void region employed, and the solid black line denotes the trajectories of the UAV.}
    \label{fig:wombat_trial_example}    
\end{figure}

Table.\ref{tab:wombat_table} presents a quantitative summary of the results of our field experiments. Five missions were carried out to localize two radio collars found to be detectable from Wombat dwelling underground. With the exception of the last mission, we were able to localize both wombats within an average of $\SI{252}{\second}$ with a mean localization error of $\SI{40}{\meter}$.  
In mission~5, after $\SI{200}{\second}$, the signal from wombat $2$ could no longer be detected before it was localized. We suspect that the wombat moved deeper underground, resulting in further signal attenuation. The intermediate belief density (particle distributions) from mission~1 are illustrated in Figure~\ref{fig:wombat_trial_example} and demonstrate the effectiveness of the information-theoretic planning objective   to reduce the uncertainty of the estimated location of Wombats.

\begin{figure}[hbt!]
    \centering
    \includegraphics[width=0.9\columnwidth]{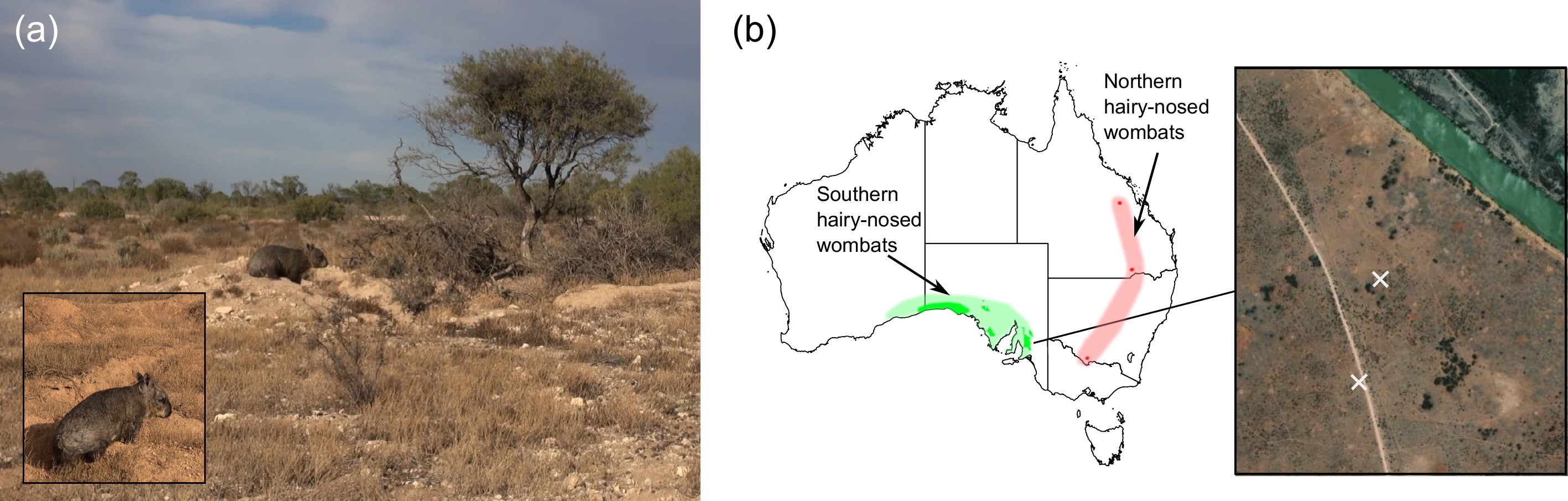}
    \caption{(a) Photograph showing a southern hairy-nosed wombat captured and released after radio tagging into the habitat at Swan Reach in Australia. (b) Main figure: Current (dark color) and extrapolated (light color) distribution of hairy-nosed wombats---based on Figure~1 in \parencite{wombat_distro}. Inset: map of the terrain at Swan Reach in Australia. The ground truth location of tagged wombats is marked by '$\times$'.}
    \label{fig:wombat_distro}
\end{figure}

\begin{table}[hbt!]
    \centering
    \begin{threeparttable}
    \caption{Results from field trials where the ConservationBot was deployed to track and locate underground radio-tagged southern hairy-nosed wombats in a conservation project.}
    \small
    \begin{tabular}{lcccc}
        \toprule
         & \multicolumn{3}{c}{Error (m)} & Time (s) \\
          \cmidrule(lr){2-4}   \cmidrule(lr){5-5}
          & Wombat 1 & Wombat 2& Mean &  Total \\
         \midrule
         Mission 1 & 48.4 & 8.0 & 28.2& 203 \\
         Mission 2 & 43.4 & 62.2& 52.8& 291 \\
         Mission 3 & 49.4 & 25.7& 37.5& 231 \\
         Mission 4 & 31.5 & 52.6& 42.1& 281 \\
         Mission 5\tnote{\dag} & 35.1 & 84.9&  -  & 200 \\
        \midrule
        Mean\tnote{\ddag} & 43.2 & 37.1 & 43.3 & 252 \\
        \bottomrule
    \end{tabular}
    \begin{tablenotes}
    \item[\dag] Mission did not locate Wombat~2 due to loss of detections from Wombat~2's VHF radio collar tag, potentially as a result of the Wombat moving deeper underground during the trial.
    \item[\ddag] Excluding mission 5
    \end{tablenotes}
    \label{tab:wombat_table}
    \end{threeparttable}
\end{table}

\subsection{Ethics and Regulatory Compliance}
This study was conducted under the University of Adelaide Animal Ethics permit number S-2018-112a. All of the flights were undertaken with the Civil Aviation Safety Authority (CASA, Australia) approvals and followed the safety protocols mandated by The University of Adelaide as such our experiments were designed around the University of Adelaide and CASA regulations governing the conduct of UAV research.  The two pilots conducting and supervising the experiments had a Remote Pilot License (RePL).

\section{Lessons Learned} \label{sec:lessons_learned}
This study describes, for the first time, the development and optimization of an aerial robotic system for the tracking and monitoring of wildlife across a variety of vegetation and terrain conditions.  We present innovative solutions that address practical and technical issues impacting radio tag detection---speed, accuracy, and reliability, and evaluate the proposed algorithms, their integration, and operation in field conditions.
In this section, we reflect upon our observations and lessons learned during our extensive field experiments as well as potential future work.

Given the significant improvements to the detection range of the software-defined VHF receiver, the maximum search area of our system is primarily limited by its flight time. For a given UAV platform, flight time is dependent on the weight of the payload. The total mass of the payload of the sensing and computing hardware is $\SI{550}{\gram}$, where the antenna we employed contributes to more than $45\%$ of the total payload. To reduce the mass of the payload, and increase flight times, a customized directional antenna design with lighter materials can be investigated in further research.

Our software systems onboard the UAV employed the existing $\SI{915}{\mega\hertz}$ telemetry channel  used for communication between the UAV and the ground station to provide data for the localization and planning algorithm  executed on the ground control station as well as the transmission of control action to the UAV. The choice of $\SI{915}{\mega\hertz}$ provided a superior range compared to the $\SI{2.4}{\giga\hertz}$ wireless link used in~\parencite{hoa2019jofr} and removed the need for an additional transceiver onboard the UAV. We found the exploitation of the telemetry channel to be convenient and use the full capability to monitor the UAV operations and meet regulatory compliance requirements with the benefit of being able to use open-source hardware and software to support the development of the robotic platform. But, we observed a packet drop rate of around 10\%, predominantly due to the limited communication channel quality. Although executing the algorithms  onboard the  companion computer for better reliability and ease of use addressed the problem, for safety reasons, in the testing phase, we could not employ this mode of operation.

The software-defined radio receiver design allows us to easily facilitate the simultaneous detection of multiple radio collar signals at different frequencies whilst facilitating the realization of the receiver in a small form factor and a lightweight hardware realization. The software programmable hardware simplifies the reconfiguration on the fly, such as receiver SNR and radio tag frequencies. Despite the advances made compared to previous software-defined receiver designs \parencite{hoa2019jofr} to increase the scanning range, tracking underground radio-collared wildlife was a challenging proposition. For our detector, a $\SI{-70}{\dBm}$ SNR threshold is used to minimize false detections (false alarms); this setting achieved a scanning range of over $\SI{2}{\kilo\meter}$. However, as shown in \ref{sec:wombat_trial}, detecting very weak signals from underground VHF radio emitters due to significant signal attenuation through the soil was challenging in this setting. 

Reducing the detector threshold could increase the probability of weak signals being detected, but it will also increase the probability of receiving false alarms. Notably, our current detector implementation only reports up to one peak detection per radio tag transmit period; whenever a false alarm is reported, the truth signal (if present) will be suppressed, which also effectively reduces the detection probability. To allow using a lower SNR to improve the detection range further, the detector architecture can be modified to report all signal detection peaks (from the peak detection stage in our detector). This will allow us to fully utilize the Bernoulli filter's ability to handle object state estimation in the presence of multiple false alarms in addition to missed detections and, therefore, increase the capability of our estimation algorithm to function under increased false alarms. Consequently, the operating range of our ConservationBots will effectively increase---more importantly, enable the tracking and localization in the presence of weak VHF radio collar tag signals, such as those from underground dwelling animals.

\section{Conclusion} \label{sec:conclusion}
We have validated the capability of our proposed approach to rapidly localize multiple mobile objects in different environments through extensive simulation-based experiments and field experiments with a prototype robotic platform---ConservationBot. Further, we have shown that our approach, which utilizes both RSSI and AoA measurements and performs measurement and trajectory planning to locate radio-collared wildlife, delivers consistently fast, robust, and better performance over traditional RSSI-only or AoA-based approaches, even when the proposed imprecise RSSI model formulation and compensated AoA detector are employed with previous approaches. Importantly, the ability to plan for measurements allows the robot to benefit from robust AoA measurements without the impact of the increased time needed to complete a mission. Although the use of the imprecise likelihood method is not a perfect replacement for the precise and correct likelihood function, it greatly simplifies the difficult task of modeling and building the often complex likelihood model significantly impacted by environmental conditions.  

Our field experiments confirm that autonomous aerial robots capable of fast, robust tracking of multiple wildlife can provide benefits over the labor-intensive manual tasking to gather precise information from wildlife for their conservation and management.

\printbibliography

\end{document}